\documentclass[lettersize,journal]{IEEEtran}
        
\usepackage{color,amsmath,amssymb,amsthm,epsfig,mathrsfs,cite,bm,graphics}
\usepackage{algpseudocode}
\usepackage{algorithmicx}
\usepackage{algorithm} 
\usepackage[utf8]{inputenc}
\usepackage{cite}
\usepackage{amsmath}

\usepackage{amsfonts}
\usepackage{amssymb}
\usepackage{placeins}
\usepackage{graphicx}
\usepackage{bbm,bm}
\usepackage{graphicx,graphics,subcaption}
\usepackage{tikz}
\usepackage{pgf}
\usepackage{pgfplots}
\usepackage{hyperref}
\usepackage[all]{hypcap} 
\usepackage{float}
\usepackage{tabularx}

\usetikzlibrary{automata, positioning, arrows,shapes.multipart,fit}

\usepackage{todonotes}
\presetkeys{todonotes}{color=blue!20}{}
\usepackage{pgfplots}
\usepackage{pgfplotstable}
\usepackage{booktabs}
\usepackage{multirow}

\usepackage{amssymb}
\usepackage{pifont}
\newcommand{\xmark}{\ding{55}}%

\title{First line of defense: A robust first layer mitigates adversarial attacks}
\author{Janani Suresh, Nancy Nayak, and Sheetal Kalyani\\
\thanks{Janani Suresh and Sheetal Kalyani are with the Department of Electrical Engineering, Indian Institute of Technology Madras, India. Nancy Nayak is with Department of Electrical and
Electronic Engineering, Imperial College London, London, UK. Emails: ee22s079@smail.iitm.ac.in, \, n.nayak@imperial.ac.uk,\,
skalyani@ee.iitm.ac.in}
}
\usepackage{bibentry}

\begin{document}

\maketitle

\begin{abstract}
      Adversarial training (AT) incurs significant computational overhead, leading to growing interest in designing inherently robust architectures. We demonstrate that a carefully designed first layer of the neural network can serve as an implicit adversarial noise filter (ANF). This filter is created using a combination of large kernel size, increased convolution filters, and a maxpool operation. We show that integrating this filter as the first layer in architectures such as ResNet, VGG, and EfficientNet results in adversarially robust networks. Our approach achieves higher adversarial accuracies than existing natively robust architectures without AT and is competitive with adversarial-trained architectures across a wide range of datasets. Supporting our findings, we show that (a) the decision regions for our method have better margins, (b) the visualized loss surfaces are smoother, (c) the modified peak signal-to-noise ratio (mPSNR) values at the output of the ANF are higher, (d) high-frequency components are more attenuated, and (e) architectures incorporating ANF exhibit better denoising in Gaussian noise compared to baseline architectures. Code for all our experiments are available at \url{https://github.com/janani-suresh-97/first-line-defence.git}.
\end{abstract}

\section{Introduction}

In adversarial attacks, a subtle perturbation to the input can cause the model to make erroneous predictions, leading to a significant drop in accuracies \cite{goodfellow2014explaining}. This necessitates robust defense mechanisms, which have garnered significant research interest \cite{chakraborty2021survey}. The white-box attacks like Fast Gradient Sign Method (FGSM) \cite{goodfellow2014explaining}, Projected Gradient Descent (PGD) \cite{madry2017towards}, and Auto Attack (AA) \cite{croce2020reliable} have full access to model architecture and parameters which leads to significant performance loss. Several defense methods have been proposed to obtain robust Deep Neural Networks (DNN) \cite{s.2018stochastic,xie2017mitigating} starting from adversarial training (AT) \cite{madry2017towards} to data preprocessing techniques \cite{zhang2019theoretically}. In \cite{yan2021cifs}, the authors improve the adversarial robustness of convolutional neural networks by channel-wise importance-based feature selection during AT. Randomized defense mechanisms, which involve the introduction of various perturbations during the inference phase to disrupt adversarial attacks, have also garnered great interest \cite{ma2024adversarial}.

AT is widely used among different defense methods, and it primarily focuses on improving adversarial robustness by training the network architectures with adversarial samples. AT is computationally intensive because iterative algorithms like PGD or AA require multiple gradient computations per training iteration. In the case of AA, the algorithm uses a combination of different attack techniques (e.g., extended version of PGD, white-box,  and decision-based attacks) to generate a wide range of adversarial examples. Running multiple attacks increases computation time as one needs to execute several different adversarial generation processes.

However, there is still a notable gap in understanding the influence of architectural components on adversarial robustness compared to the attention given to AT methods. The authors in \cite{huang2021exploring} investigate the impact of architectural elements such as topology, depth, and network width on the robustness of adversarially trained DNNs with a focus on ResNets to enhance the adversarial robustness. To circumvent the complexity associated with AT, \cite{lukasik2023improving} investigate native robustness. They show that while the models trained with clean samples often prioritize high-frequency information, AT shifts focus from high-frequency to low-frequency details. By leveraging this fact, the authors enhance the native robustness through the regularization of filters with high frequencies. We show that one can achieve native robustness by designing an \textit{Adversarial Noise Filter} (ANF) that inhibits the passage of adversarial noise, thereby achieving a robust neural network by using only our proposed ANF layer as the very first layer. The authors in \cite{xie2019feature} have also used network blocks to denoise the features via AT; however, unlike their work, we focus on native robustness and modify only the first layer of the architecture to denoise the features.

In designing the ANF, which serves as the first layer of the neural network, we employ a combination of three operations: (a) increased kernel size, (b) higher number of filters, and (c) maxpool, and show that solely by utilizing these operations, one can achieve a substantial degree of robustness without using AT. We hypothesize and then show through extensive empirical results that the proposed combination of kernel, filters, and maxpool implicitly filters out the adversarial noise and reduces its propagation to other layers. 

We provide a detailed discussion on ANF and incorporate it as the first layer in a variety of popular architectures. We then show that a robust first layer such as ours behaves like an adversarial noise filter/denoiser by analyzing a metric closely related to PSNR, which we call the mPSNR (modified Peak Signal-to-Noise Ratio). We show that mPSNR substantially improves once we pass the adversarially perturbed image through the proposed ANF-based first layer. In other words, the ANF serves as a first layer of defense and is sufficient to mitigate adversarial noise. Since our method is natively robust, we compare it with one of the most recent works in this area by \cite{lukasik2023improving} and the other native robustness methods cited therein. Note that all existing works on native robustness typically change all/multiple layers of the neural network to be natively robust \cite{lukasik2023improving,grabinski2022frequencylowcut,lopes2019improving}. Ours is the first work to show that native robustness can be achieved by only changing the first layer of the architecture. Furthermore, we achieve substantially better robust accuracies than state-of-the-art (SOTA) native robust architectures despite changing only the very first layer. To further demonstrate the practical utility of our work, we design one single ANF/first layer and then incorporate that as the first layer for a variety of architectures, i.e., we do not change the number of kernels, filters, or maxpool operators across the architectures. Nevertheless, we observe substantially higher adversarial accuracies across architectures and attacks. For example, we are $15.56\%$ and $26\%$ better than \cite{lukasik2023improving} against PGD and AA for ResNet20 on CIFAR10. 
We show results for FGSM, PGD, and AA on VGG, ResNets, WideResNets, and EfficientNet architectures across datasets such as CIFAR10, CIFAR100, TinyImagenet, and Imagenet.

To further investigate and compare our work with the corresponding baseline architectures, we study the loss surfaces and the contour plots inspired by \cite{li2018visualizing} and observe that our method smoothens the adversarial loss surface and achieves lower adversarial loss. Moreover, utilizing the decision boundary visualization technique from \cite{schulz2019deepview}, we study the change in decision boundaries with and without ANF, and this further showcases the robustness of our approach. Aside from the visible robustness demonstrated using the above two visualizations, we demonstrate the robustness of our method to high variance Gaussian noise. Further, we study the spectrum of adversarial noise when passed through ANF and show that ANF attenuates higher frequencies, which is what AT typically tries to achieve \cite{gilmer2019adversarial} and we also show how the proposed method acts as a feature denoiser.

\section{Proposed technique}
Drawing inspiration from signal processing research, where one combats noise by designing filters resilient to noise, we propose combining larger kernels, a higher number of filters, and a maxpool operator to construct ANF. We use ANF as the first layer of the network architecture, and this restricts the passage of noise to the other layers.

\subsection{Basic components of ANF}
\textbf{Kernel size: }
Typically $3\times 3$, $5\times 5$, $7 \times 7$ are popular choice for kernel size \cite{ozturk2018convolution}. However, recently \cite{Ding_2022_CVPR} have utilized sizes as large as $31\times 31$ to achieve better performance with convolutional neural networks because large kernels have larger receptive fields, which implies an increase in information to subsequent convolution blocks. The larger kernels also allow the network to capture more global patterns in the input data, which can benefit tasks where context matters, such as image classification. Denoising processes also benefit from using larger kernels, as incorporating and averaging more pixels effectively reduces variance \cite{vogels2018denoising}. We leverage that larger kernels smooth the features and hence should also smooth/mitigate noise. Inspired by \cite{Ding_2022_CVPR} and \cite{vogels2018denoising}, the existing kernel dimension is increased from $k\times k$ to $\tilde{k}\times\tilde{k}$ only for the ANF layer. Replacing every small kernel with a corresponding large kernel is prohibitively expensive. Fortunately, in our work, we change only the first layer; hence, the associated computational cost is less.

\textbf{Increased number of filters: }
More filters allow the network to learn a diverse set of features at each layer, which helps the network generalize better to different types of inputs and variations in the data, making it more robust to changes in the inputs. Let the number of filters in the baseline be $F$, then the number of filters in the first convolution layer (i.e., ANF) is increased to $\tilde{F}$.

\textbf{Maxpool:}
Maxpool is a downsampling operation that reduces the spatial dimensions of the feature map while retaining the essential information. It is applied after convolution to reduce computational complexity and control overfitting. Maxpool divides the input into non-overlapping rectangular regions and outputs the maximum value from each region. This reduces the size of the feature map while preserving the most prominent features. Maxpool serves as a down-sampling operator, and hence, we hypothesize that it can downsample or reduce the impact of adversarial noise. See Sec. \ref{sec:investigation}, where we demonstrate how maxpool is crucial for the ANF design.

To show that the combination of the above three operations leads to denoising, we measure the denoising capability of the ANF by calculating the noise in the feature before and after the ANF.

\subsection{Measure of denoising}
Peak Signal-to-Noise Ratio (PSNR) is a metric that quantifies the ratio between the maximum possible power of the original image and the power of noise that corrupts the image, thereby assessing the fidelity of the reconstructed image. 
As the dimensions of all the input samples are the same, we denote the dimension of sample $\mathbf{x}_i$ by $[W, H, D]$ where $D$ is the number of features in $\mathbf{x}_i$; $W$ and $H$ are the width and height of a single input feature $\mathbf{x}_i^{d}$. Let the perturbed image be denoted as $\tilde{\mathbf{x}}_i$. The transformed feature for $\mathbf{x}_i$ and $\tilde{\mathbf{x}}_i$ after the ANF is denoted by $\mathbf{h}_i$ and $\tilde{\mathbf{h}}_i$, respectively. 

The PSNR is typically computed for one particular input feature. For the adversarial attacks, all $N$ features may not always have non-zero values for the difference between the actual feature $\mathbf{h}_i$ and the corrupted feature $\tilde{\mathbf{h}}_i$. This results in zero in the denominator if one uses PSNR. To handle such a situation, we first compute the inverse of the PSNR as shown below for every sample, where the numerator is a measure of the perturbation in the feature, and the denominator is the peak value of the input image:
\begin{align}
    \text{invPSNR}_i = \frac{\sqrt{\frac{1}{DHW}\sum_{d=1}^{D}\sum_{h=1}^{H}\sum_{w=1}^{W} (h_{idhw}-\tilde{h}_{idhw})^2}}{\frac{1}{D}\sum_{d=1}^{D} \max_{1\leq m \leq H, \, 1\leq n \leq D}^{} \,\mathbf{x}_{imn}^{d}}.
    \label{eq:invPSNR}
\end{align}
This is followed by an average over $N$ samples and then taking an inverse as follows:
\begin{align}
    \text{mPSNR} &= \frac{1}{\frac{1}{N}\sum_{i=1}^{N} \text{invPSNR}_i}
\end{align}
We define the above metric as modified PSNR (mPSNR) and measure it at the input layer and after the ANF to understand how well the ANF can denoise the perturbed image. In the next section, we briefly describe various adversarial attacks.

\section{Adversarial attacks}
Let $\mathbf{x}_i\in \mathbb{R}^d$ be an input data point belonging to class $c_o$, and $y_i$ is the corresponding true label. Let the loss function be given by $L(\mathbf{x}, y; \mathbf{w})$, where $\mathbf{x}$ is the input, $y$ is the corresponding true label, and $\mathbf{w}$ is the network parameters. An adversarial attack is a mapping $\mathcal{A}: \mathbb{R}^d \rightarrow \mathbb{R}^d$ such that the perturbed data $\tilde{\mathbf{x}}_i=\mathcal{A}(\mathbf{x}_i)$ is misclassified as an adversarial class $c_t$. We focus on white-box attacks, which assume full knowledge of the classifier and require knowing all weight values and the network structure. We show the results for three popular white-box attacks, i.e., FGSM, PGD, and AA.

\begin{figure}[!t]
    \centering
    \includegraphics[scale=0.42]{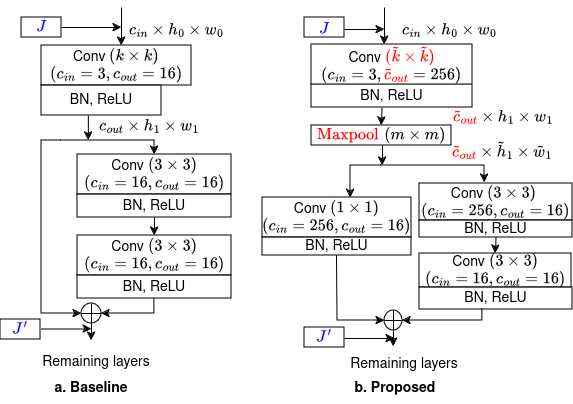}
    \caption{ANF with ResNet20}
    \label{fig:resnets}
\end{figure}

\textbf{FGSM: } The loss function is maximized, subject to an upper bound on the input perturbation, i.e., $||\tilde{\mathbf{x}}_i-\mathbf{x}_I ||_\infty \leq \epsilon$. The FGSM creates an attack $\tilde{\mathbf{x}}_i$ by
\begin{equation}
    \tilde{\mathbf{x}}_i = \mathbf{x}_i + \epsilon\, \operatorname{sign}(\nabla_{\mathbf{x}}L(\mathbf{x}_i, y_i; \mathbf{w})).
\end{equation}
Here, $\nabla_{\mathbf{x}}L(\mathbf{x}_i, y_i; \mathbf{w})$ is interpreted as the gradient of $L$ with respect to $\mathbf{x}$ evaluated at $\mathbf{x}_i$.

\textbf{PGD: }
PGD iteratively perturbs input data to maximize the model's loss $L(.; \mathbf{w})$ to find more effective adversarial examples compared to non-iterative attacks. The adversarial sample is found in the same way as FGSM but iteratively starting from $\tilde{\mathbf{x}}^{(0)} = \mathbf{x}_i$, one updates the sample according to the following:
\begin{align}
     \tilde{\mathbf{x}}^{(k+1)} = \text{Proj}_{B_{\xi}^{} (\mathbf{x}_i)}\left(\tilde{\mathbf{x}}^{(k)} + \eta \,\, \operatorname{sign}\left( \nabla_{\mathbf{x}}L(\tilde{\mathbf{x}}^{(k)},y_i; \mathbf{w})\right)\right) 
\end{align}
for $k=0, 1, \dots, K-1$, where $K$ is the number of PGD iterations, $\eta$ is the step size, $\xi$ is the norm-ball radius, and $\text{Proj}_{B_{\xi} (\mathbf{x})}(\mathbf{z})= \arg \min_{\mathbf{ x}' \in B_{\xi}(\mathbf{ x})} \|{\mathbf{z}-\mathbf{x}'}\|_p$ \cite{madry2017towards}. The final adversarial sample is $\tilde{\mathbf{x}}_i=\tilde{\mathbf{x}}^{(K)}$. The PGD attack has been widely used to evaluate machine learning models' robustness and develop more effective defense mechanisms against adversarial attacks \cite{huang2023revisiting}. 

\textbf{AA: } In Auto attacks, which is one of the most powerful attacks, two different versions of PGD i.e. AutoPGD with Cross Entropy loss and Difference of Logits Ratio loss proposed in \cite{croce2020reliable} are combined with white-box FAB attack \cite{croce2020minimally}, and black-box Square Attack \cite{andriushchenko2020square} to form a parameter-free ensemble of complementary attacks. For more details on AA, please refer to \cite{croce2020reliable}. 

\section{Results and Discussions}
\label{sec:resultsanddiscussion}
We now discuss how ANF is incorporated as the first layer of ResNet, WideResNet, VGG, and EfficientNet architectures based on their structures\footnote{The baseline performance for each architecture has been sourced from the following references; \\\textbf{ResNet20}: \url{https://github.com/akamaster/pytorch_resnet_cifar10}, \\\textbf{WRN-28-10}: \url{https://github.com/bmsookim/wide-resnet.pytorch}, \\\textbf{VGG16 and ResNet50}: \url{https://github.com/kuangliu/pytorch-cifar/blob/master/models/resnet.py}, \\\textbf{EfficientNet-B0}: \url{https://github.com/jovitalukasik/filter_freq_reg}}. For comparison with very recent work on native robustness \cite{lukasik2023improving}, we showcase results on the same baselines, i.e., ResNet20 and EfficientNet-B0. In line with \cite{lukasik2023improving}, the attack strength for all the attacks is considered to be $\epsilon=1/255$, and the number of iterations for the PGD attack is considered to be $40$, unless specified otherwise. We have used $\ell_{\infty}$ norm for AA apart from the default hyper-parameters\footnote{\url{https://github.com/fra31/auto-attack}.}.

\subsection{ANF with ResNet and WRN}
The ANF in ResNet20 for CIFAR10 and CIFAR100 classification is represented by Fig. \ref{fig:resnets}, where the differences between the baseline and the proposed are highlighted in red. The kernel size is increased from $k\times k$ to $\tilde{k}\times \tilde{k}$, where $k=3$, and $\tilde{k}=15$. The number of output features from the first convolution layer is increased from $F=c_{out}=16$ to $\tilde{F}=\tilde{c}_{out}=256$. Further, the maxpool is introduced with a kernel size of $m=5$ and a stride of $1$.

\begin{table}[!t]
    \centering
    \footnotesize 
    \setlength{\tabcolsep}{3mm} 
    \begin{tabular}{p{3.0cm}p{0.5cm}p{0.5cm}p{0.5cm}p{0.5cm}p{0.5cm}}
    \toprule
        Arch.  & FGSM & PGD & AA & CorptnA & Clean acc\\
    \midrule
        \multicolumn{6}{c}{ResNet20 with CIFAR10} \\
    \midrule    
        Baseline & $42.86$ & $27.03$ & $12.41$ & $73.32$ & $\textbf{91.26}$ \\
        ANF & $\textbf{59.56}$ & $\textbf{59.98}$ & $\textbf{55.14}$ & $\textbf{78.43}$ & $83.09$ \\
        \cite{lukasik2023improving} &  $53.12$ & $44.42$ & $29.14$ & $-$ & $90.54$ \\
        AT \cite{lukasik2023improving} & $49.93$ & $46.34$ & $36.47$ & $-$ & $70.31$ \\
        FLP \cite{grabinski2022frequencylowcut} & $52.49$ & $30.25$ &  $8.48$ & $-$ & $91.52$ \\
        GA \cite{lopes2019improving} &  $50.36$ & $31.50$& $11.38$& $-$ & $91.29$ \\
    \toprule
        \multicolumn{6}{c}{ResNet20 with CIFAR100} \\
    \midrule
        Baseline  & $12.28$ & $3.83$ & $1.01$ & $34.93$ & $\textbf{65.34}$ \\
        ANF & $\textbf{26.8}$ & $\textbf{26.43}$ & $\textbf{21.58}$ & $\textbf{48.13}$ & $54.86$ \\
        \cite{lukasik2023improving} &  $17.2$ & $12.24$ & $5.11$ &  $-$ & $58.19$ \\
    \toprule
        \multicolumn{6}{c}{WRN-28-10 with CIFAR100} \\
    \midrule
        Baseline & $29.51$ & $20.67$ & $10.88$ & $51.1$ & $\textbf{77.9}$ \\
        ANF &  $\textbf{43.8}$ & $\textbf{44.38}$ & $\textbf{41.09}$ & $\textbf{58.24}$ & $62.22$ \\
    \toprule
        \multicolumn{6}{c}{VGG16 with CIFAR10} \\
    \midrule
        Baseline & $60.60$ & $54.94$ & $43.83$ & $\textbf{75.06}$ & $\textbf{92.42}$ \\
        ANF & $\textbf{62.27}$ & $\textbf{63.11}$ & $\textbf{60.42}$ & $67.14$ & $80.44$ \\
    \toprule
        \multicolumn{6}{c}{EfficientNet-B0 with CIFAR10} \\
    \midrule
        Baseline  & $53.05$ & $52.20$ & $42.24$ & $44.08$ & $\textbf{92.29}$ \\
        ANF  & $\textbf{64.95}$ & $\textbf{66.23}$ & $\textbf{62.27}$ &  $\textbf{80.18}$ & $87.14$ \\
        \cite{lukasik2023improving} & $57.83$ & $59.68$ & $53.50$ & $-$ & $89.18$ \\
        \bottomrule
    \end{tabular}  
    \caption{Comparison of different architectures with CIFAR10 and CIFAR100. The highest accuracies are highlighted.}
    \label{tab:combined}  
\end{table}

\begin{table}[!t]
    \centering
    \footnotesize 
    \setlength{\tabcolsep}{3mm} 
    \begin{tabular}{p{3.0cm}p{0.6cm}p{0.6cm}p{0.6cm}p{0.6cm}}
    \toprule
        Arch. & FGSM & PGD & AA & Clean acc\\
    \midrule
        \multicolumn{5}{c}{ResNet50 with TinyImagenet} \\
    \midrule 
        Baseline & $34.07$ & $31.82$ & $24.08$ &  $\textbf{69.7}$ \\
        ANF &  $\textbf{35.64}$ & $\textbf{35.17}$ & $\textbf{28.08}$ & $60.98$ \\
    \toprule
        \multicolumn{5}{c}{ResNet50 with ImageNet} \\
    \midrule
        Baseline  & $42.36$ & $26.17$ & $1.05$ & $\textbf{64.37}$ \\
        ANF with AT &  $\textbf{55.09}$ & $\textbf{55.46}$ & $\textbf{ 52.95}$ & $61.67$ \\
        AT \cite{lukasik2023improving} & $36$ & $37$ & $24.32$ &$58.09$ \\
    \bottomrule
    \end{tabular}
    \caption{Comparison of ResNet50 with TinyImagenet and ImageNet.The highest accuracies are highlighted.}
    \label{tab:resnet50}
\end{table}
The ANF in ResNet50 used for TinyImageNet classification includes some modifications to the first layer of the baseline architecture compared to ResNet20. For ResNet50, a key difference in the first layer is that the number of filters is $64$ instead of $16$. In our proposed architectures, we increase the filters to $256$, the kernel size increased from $3 \times 3$ to $15 \times 15$, and use a max-pooling operation with a $5 \times 5×$ dimension. The WRN considered in our work for CIFAR100 classification has pre-activation, which implies that the batch normalization (BN) and ReLU are before the convolution in every basic block. Otherwise, the first layer transformation is just the same as ResNet20.

In ResNets and WRNs, a common architectural characteristic includes residual blocks, typically positioned after the first convolutional layer. We convert this first convolution layer to ANF by introducing maxpool and by increasing the kernel size and the number of filters. To explain it further, the image samples in CIFAR10, CIFAR100, TinyImagenet, and Imagenet datasets have three features, e.g., `R', `G', and `B'; therefore $c_{in}=3$. For the ResNet and WRN architectures we have chosen, the number of output features from the first convolution layer is $16$, which means $c_{out}=16$. Note that, with the proposed ANF for the ResNet and WRN architectures, the number of features input to the first residual block changes from $16$ to $256$ as we increase the number of filters from $16$ to $256$ at the first convolution layer. So, to match the dimensions in the succeeding layers, the signal in the direct path is passed via a $1\times 1$ convolution before summing it up with the features from the residual path. The usual ResNet architectures already have this $1\times 1$ convolution, which matches the dimensions before adding every residual block.

From Table \ref{tab:combined}, for ResNet20 with CIFAR10 dataset, the proposed ANF achieves PGD and AA accuracy of $59.98\%$ and $55.14\%$ respectively compared to $27.03\%$ and $12.41\%$ in the baseline. By using frequency filter regularization \cite{lukasik2023improving} show that their method outperforms frequencylowcut-pooling (FLP) \cite{grabinski2022frequencylowcut} and patch Gaussian augmentation (GA) \cite{lopes2019improving} by a significant margin in terms of native robustness. AT \cite{lukasik2023improving} in Table \ref{tab:combined} represents AT results that the authors have produced by training their architecture with FGSM attacks with the attack strength of $\epsilon=8/255$ for CIFAR10. However, our proposed ANF significantly outperforms both \cite{lukasik2023improving} and AT in \cite{lukasik2023improving} without using AT. We also evaluate the proposed ANF against corruption noise, which is Gaussian noise with a standard deviation of $\sigma=16/255$. We achieve an accuracy of $78.43\%$ against corruption noise with the help of ANF, whereas the baseline has an accuracy of $73.32\%$. A model that has better adversarial accuracy often suffers from clean accuracy. However, with ANF, one does not have to perform AT, and therefore, we can maintain a clean accuracy of $83.09\%$ compared to $70.31$ with AT. Our approach achieves an accuracy of $21.58\%$ compared to $5.11\%$ reported by \cite{lukasik2023improving} for ResNet20 on CIFAR100 for AA, a powerful attack. For all ResNet variants and WRN-28-10, we observe significant accuracy improvement for all attacks and across datasets in Tables 1 and 2 when compared with the corresponding baselines and \cite{lukasik2023improving}. 

\textbf{Effect of maxpool stride:} In the ANF filter, the maxpool operation has stride of one. We have also considered maxpool with stride of two and given results in subSec. C of the supplementary section. Coarser downsampling with stride two provides even better adversarial accuracy than stride one. Further, the performance of ResNet20 is tested with attacks of different strengths and given in subSec. C of the supplementary section. 

\textbf{ImageNet dataset: }Models trained with ImageNet dataset barely withstand adversarial attacks \cite{lukasik2023improving} and even ANF alone cannot provide native robustness. However, ANF provides significant robustness when trained with adversarial samples. We use ResNet50, where the variant designed for ImageNet has a kernel size of $7 \times 7$ in the first layer, $64$ filters, and a max-pooling kernel size of $3 \times 3$. Our proposed architecture with ANF modifies these parameters to a kernel size of $15 \times 15$, $256$ filters, and a max-pooling kernel size of $5 \times 5$ in the first layer. In line with the work by \cite{lukasik2023improving}, who performed AT on ImageNet using ResNet50, we also trained the ANF incorporated ResNet50 adversarially with a 1-step PGD attack at $\epsilon = 4/255$. Our results demonstrate a significant improvement in accuracy achieving $55.09\%$, $55.46\%$, and $52.95\%$, compared to $36\%$, $37\%$ and $24.32\%$ of AT \cite{lukasik2023improving} for FGSM, PGD, and AA, respectively, as shown in Table \ref{tab:resnet50}. These results were obtained using early stopping at around $200$ epochs.

\subsection{ANF with VGG}
The VGG16 architecture was proposed with multiple maxpool layers, primarily to reduce the spatial dimensions of the input feature maps, thus decreasing the computational complexity of the network. As the ANF has a maxpool layer with a kernel size of $m=5$, the spatial feature dimension is already reduced after ANF. If the remaining maxpool layers from the baseline architecture are retained, the dimensions of the features decrease significantly in the deeper layers, leading to a substantial reduction in the feature information transmitted to subsequent layers. Therefore, when we integrate ANF with VGG, we remove the other maxpool layers and use only the ANF, which includes a maxpool layer with a stride of two to keep the input feature dimensions similar to the baseline\footnote{For VGG16, we opt for a stride of $2$ for maxpool to have a similar complexity as the baseline while for other architectures the maxpool stride is 1.}. From Table \ref{tab:combined}, we observe that the robust accuracy of the VGG16 baseline is comparatively better than other baselines because of the non-linearity caused by maxpool and the usage of multiple layers of a higher number of convolution filters. This further justifies the usage of maxpool in the ANF for robustness. Compared to this baseline, the proposed architecture with ANF has the following differences. The baseline has five maxpool layers, each having a kernel size $m=2$ and a stride of $2$. For ANF, we use only one maxpool layer as a part of ANF just after the first convolution layer with kernel size $m=5$ and a stride of $2$. The number of output features for the filters at the first convolution layer is increased from $64$ to $256$. The kernel size is increased from $3$ to $15$.

With ANF, we achieve accuracies of $63.11\%$ for PGD and $60.42\%$ for AA compared to the baseline accuracies of $54.94\%$ for PGD and $43.83\%$ for AA. The resultant architecture after integrating ANF and the impact of different stride values for maxpool is shown in subSec. A in the supplementary section. 

\subsection{ANF with EfficentNet-B0}
\label{sec:effnet}
The baseline EfficientNet-B0 has $32$ filters in the first layer with a kernel size of $3\times 3$, and there is no maxpool in that layer. The proposed architecture converts the first layer to ANF with $256$ filters with a kernel size of $15\times 15$. In the ANF layer, the convolution, batchnorm, and the swish activation are followed by maxpool operation with a kernel size of $5\times 5$. Although the clean accuracy of ANF is marginally lesser than the baseline, the baseline exhibits significant vulnerability to adversarial attacks. The proposed ANF demonstrates superior robustness with minimal degradation in clean accuracy, as seen from Table \ref{tab:combined}.

\section{Why does the ANF work?}
\label{sec:investigation}
In the previous section, we have shown through extensive experiments that the ANF-based architectures exhibit significant implicit robustness to adversarial attacks without requiring AT. Further, all our architectural changes are restricted to only the first layer. In all architectures, we have chosen convolution kernel size as $15\times15$, filter number as $256$, and maxpool as $5\times5$ to show even without tuning these numbers according to the architecture, one can still get excellent adversarial robustness. In subSec. B in the supplementary section, we detail the impact of tuning K, F, and M values in the ANF for PGD, FGSM, and AA, respectively, and sometimes even better results are obtained. 

In this section, we would like to better understand the workings of ANF using various metrics and visualization tools. We focus on the ResNet20 architecture for a detailed analysis, as \cite{lukasik2023improving} extensively use ResNet20 in their study and restricts their investigation to the CIFAR10 dataset. Unless otherwise stated, the baseline and the proposed architecture with ANF are models where the data is not normalized. Note that not much change in performance was observed upon introducing data normalization. In Table \ref{tab:mPSNR}, we consider eight different scenarios for ResNet20 with combinations of three choices: (i) increased Kernel size (K), (ii) increased number of filters (F), and (iii) maxpool (M). 

\textbf{mPSNR:} To study the effectiveness of the filter to denoise the features, we study the mPSNR before and after the ANF. Just after the ANF, the dimension of the features is not the same as the baseline, as shown in Fig. \ref{fig:resnets}. Therefore, we report the mPSNR only after the first residual block when the feature dimension is the same as the baseline. We report the mPSNR at the input, i.e., at point $J$, and after the first residual block, i.e., $J'$ in Fig. \ref{fig:resnets}. The mPSNR values are calculated with all the $10$k test samples of the CIFAR10 dataset. Note that the initial mPSNR at the input, i.e., at $J$, is relatively high, as expected, as adversarial noise is a minute perturbation. However, these are carefully structured perturbations, and for them to be effective, they should lead to a drop in mPSNR after propagating through the layers of the network. This is an expected trend and is also evident in the last column of Table \ref{tab:mPSNR}. However, the drop in mPSNR is the least for our architecture, and this is proof of our work's implicit adversarial noise-denoising capability. The mPSNR at the input (at $J$) is almost in the same range for all eight versions. However, the mPSNR at $J'$ varies across different versions. ANF with only maxpool in Type 1 has an mPSNR of $65.65$, and the mPSNR increases to $78.17$ and $89.92$ by combining operations K and F with M, respectively. Further, the mPSNR is maximum when K, F, and M are combined for designing ANF. One can observe that the most robust architecture is one where the mPSNR value is maximum at $J'$. The trend in mPSNR values across different configurations corroborates the PGD accuracy as shown in Table \ref{tab:mPSNR}. 

\begin{figure*}[t]
    \centering
    \begin{subfigure}{0.35\textwidth}
    \includegraphics[width=\textwidth]{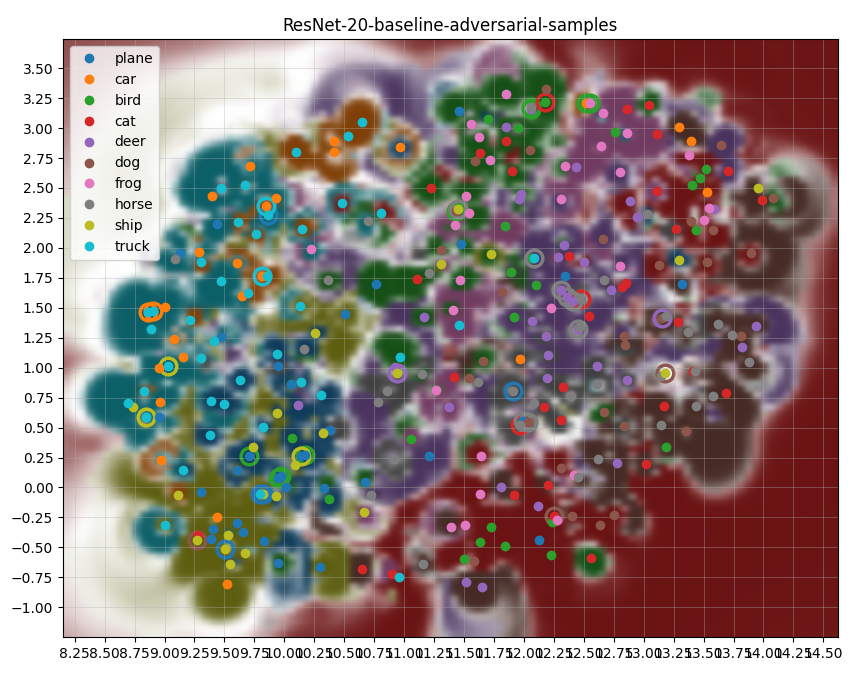}
    \caption{Baseline with adversarial samples}
    \label{fig:baseline_adv}
    \end{subfigure}
    \begin{subfigure}{0.45\textwidth}
    \includegraphics[width=\textwidth]{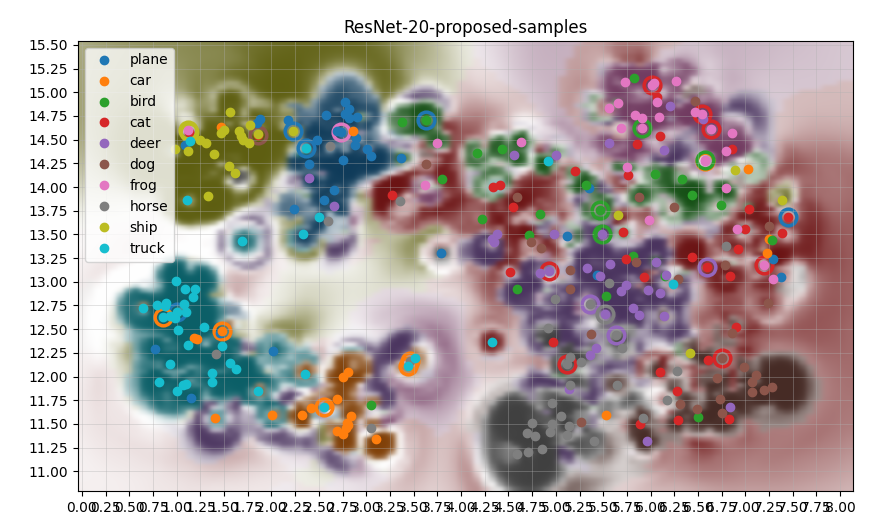}
    \caption{ANF with adversarial samples}
    \label{fig:proposed_adv}
    \end{subfigure} 
    \caption{Visualization of the decision regions}
    \label{fig:decisionregion}
\end{figure*}

We present the clean accuracy and the adversarial accuracy when tested with a PGD attack. While looking at the individual contribution of the three operations, K, F, and M in Type 1, Type 2, and Type 4, respectively, we observe that Type 1 and Type 4 have almost the same adversarial accuracy, i.e., $45.37\%$ and $45.54\%$, respectively, against PGD attack which is higher than Type 2, i.e., $29.91\%$. This indicates that among the three operations, introducing maxpool and increasing kernel size gives the highest robustness. In Type 3, when maxpool is combined with increased filter operations, it improves the robustness further to $49.92\%$. In Type 5, when kernel size is increased along with the introduction of maxpool, the adversarial accuracy improves and becomes $51.71$. Using all three operations, K, F, and M, together in Type 7 improves the accuracy to $59.93\%$, significantly higher than $27.22\%$ of the baseline. 

\begin{table}[!ht]
    \centering
    \small 
    \setlength{\tabcolsep}{3mm} 
    \begin{tabular}{p{0.9cm}p{0.2cm}p{0.2cm}p{0.2cm}p{0.5cm}p{0.5cm}p{0.6cm}p{0.6cm}}
    \toprule
        Arch. & K & F & M &  PGD  & Clean acc & mPSNR at $J$ & mPSNR at $J'$ \\
    \midrule
    Baseline & \xmark& \xmark& \xmark & $27.22$ & $91.26$ & $160.42$ & $22.66$  \\
    Type $1$ & \xmark& \xmark&  \checkmark  & $\textbf{45.37}$  & $88.35$ & $158.18$ &  $\textbf{65.65}$ \\
    Type $2$ & \xmark&  \checkmark & \xmark & $29.91$  & $91.16$ &$160.23$ &  $24.23$  \\
    Type $3$ & \xmark&  \checkmark &  \checkmark   & $\textbf{49.92}$  & $89.72$ &$156.63$ & $\textbf{61.79}$  \\
    Type $4$ &  \checkmark & \xmark& \xmark & $45.54$  & $85.68$  &$156.08$ &  $59.08$  \\
    Type $5$ &  \checkmark & \xmark&  \checkmark  & $\textbf{51.71}$  & $80.99$ & $153.06$ & $\textbf{78.17}$\\
    Type $6$ & \checkmark &  \checkmark & \xmark & $40.12$ & $86.64$ & $157.47$ & $29.80$  \\
    Type $7$ &  \checkmark &  \checkmark  & \checkmark& $\textbf{59.93}$  & $83.09$ & $151.62$ &  $\textbf{89.92}$ \\
    \bottomrule
    \end{tabular}
    \caption{mPSNR in ResNet20 for CIFAR10. For column K, \checkmark increases the kernel size from $3\times 3$ to $15\times 15$; for column F, \checkmark increases filters from 16 to 256; for column M, \checkmark introduces a $5\times 5$ maxpool operation.}
    \label{tab:mPSNR}
\end{table}

The mPSNR values clearly demonstrate that the proposed ANF effectively filters out a significant amount of adversarial noise. To better understand the impact of this filtering, we will visualize the decision surface for both the baseline and the proposed approach next.

\subsection{Visualization of the decision surface}
In the DeepView method, \cite{schulz2019deepview} propose a discriminative dimensionality reduction (DR) method called Fisher UMAP, enabling the DR to focus on the aspects of the data that are relevant to the classifier. They also develop a scheme based on inverse dimensionality reduction to obtain predictions only for a relevant subspace, which is then used to visualize the decision function in two dimensions. The steps to perform the visualization are 
\begin{enumerate}
    \item Apply Fisher UMAP which is based on the underlying deep network to project a data set consisting of points $\mathbf{x}_i$ to two dimensions, yielding $\mathbf{y}_i = \pi(\mathbf{x}_i)$ so that $\pi$ preserves the information in $\mathbf{y}_1, \dots, \mathbf{y}_n \in \mathbb{R}^d, d=2,3$ from $\mathbf{x}_1, \dots, \mathbf{x}_n \in \mathbf{S}$ as much as possible.
    \item Create a tight regular grid of samples $\mathbf{r}_i$ in 2D space and map it to high dimensional space by $\mathbf{s}_i=\pi^{-1}(\mathbf{r}_i)$
    \item Apply the neural network $f$ to $\mathbf{s}_i$ in order to obtain predictions and certainties.
    \item Visualize the label together with the entropy of the certainty for each position $\mathbf{r}_i$ in the background of the projection space in order to obtain an approximation of the decision function.
\end{enumerate}

\begin{figure*}[!ht]
\centering
\begin{subfigure}{0.245\textwidth}
    \includegraphics[width=\textwidth]{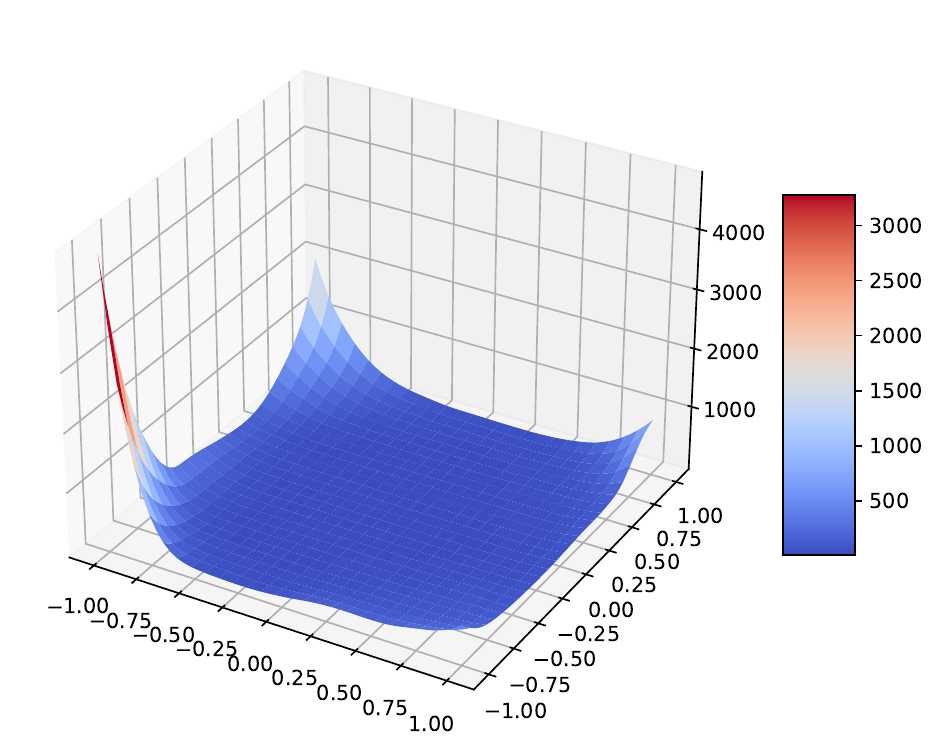}
    \caption{Baseline loss surfaces }
    \label{fig:losssurface_3d_baseline_adv}
\end{subfigure}
\begin{subfigure}{0.245\textwidth}
    \includegraphics[width=\textwidth]{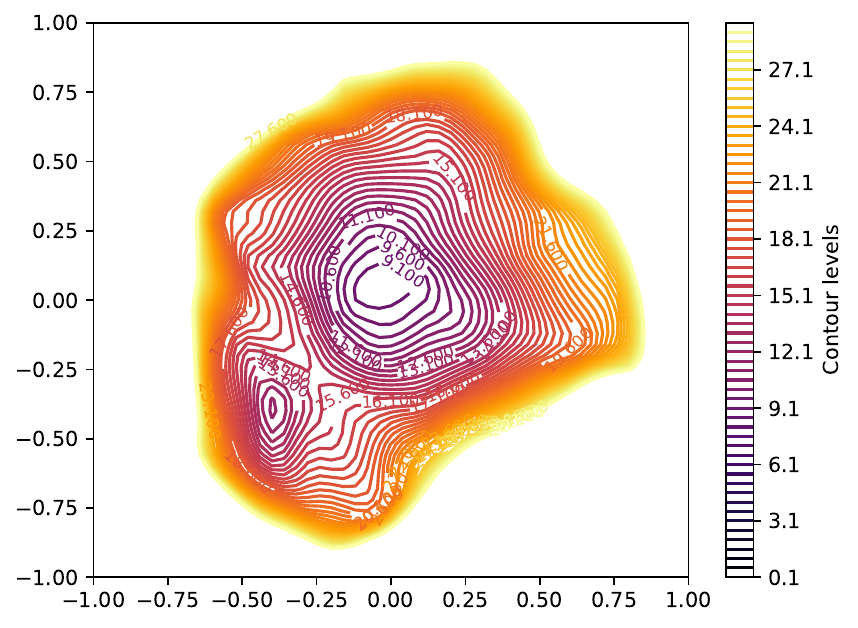}
    \caption{Baseline contour plots}
    \label{fig:losssurface_contour_baseline_adv}
\end{subfigure}
\begin{subfigure}{0.245\textwidth}
    \includegraphics[width=\textwidth]{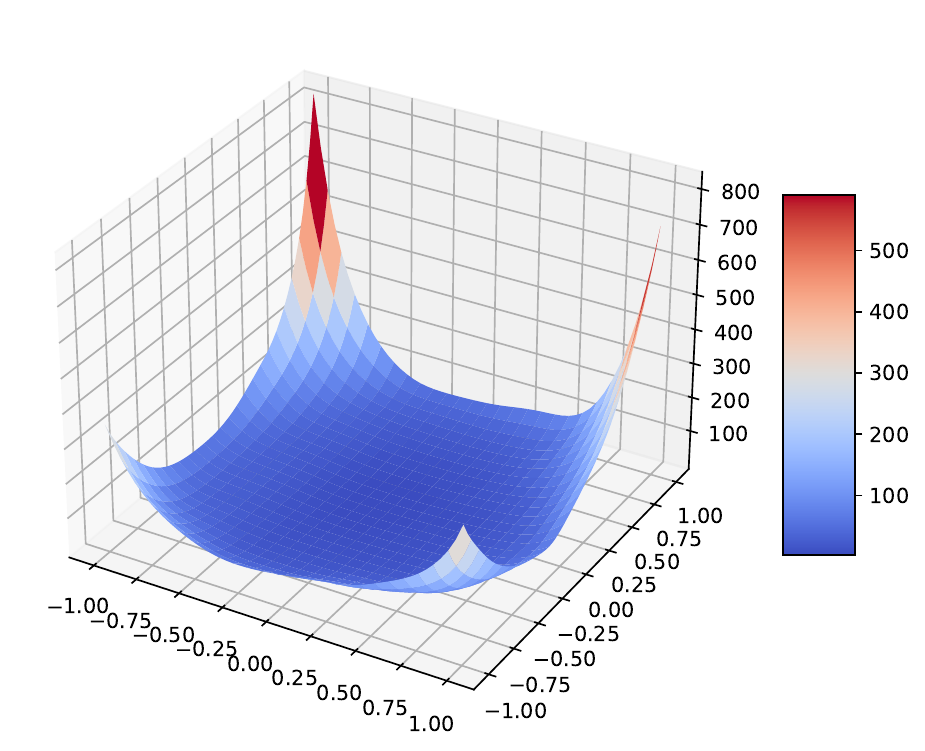}
    \caption{ANF loss surfaces}
    \label{fig:losssurface_3d_proposed_adv}
\end{subfigure}
\begin{subfigure}{0.245\textwidth}
    \includegraphics[width=\textwidth]{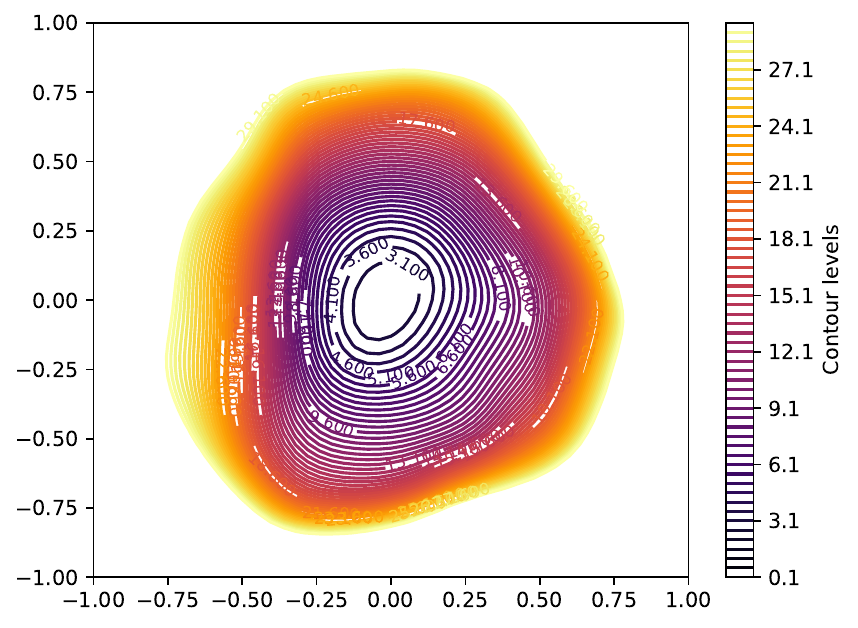}
    \caption{ANF contour plots}
    \label{fig:losssurface_contour_proposed_adv}
\end{subfigure}
\caption{Visualization of 3D loss surface (a,c) and contour plots (b, d) with adversarial samples.}
\label{fig:loss_surface}
\end{figure*}

For better representation, both baseline and proposed architecture with ANF are trained with normalized data without any change in performance. In Fig. \ref{fig:decisionregion}, Fig. \ref{fig:baseline_adv} illustrates the decision boundary of the baseline architecture using 300 adversarial samples from the test data. In comparison, Fig. \ref{fig:proposed_adv} displays the decision boundary of the proposed architecture with the same 300 samples. Each point represents an image, with its colour indicating the original label. The background colours are the predictions learned by UMAP, with intensity reflecting the confidence level of the predictions. Additionally, there are encircled representations where the outer circle of the encircled point has the colour of the class predicted by the model. The encircled points are present when the model predictions do not match the UMAP representation, i.e., the outer circle colours are different than the background colour. Among these encircled points, two cases may arise; in the first case, the model prediction is different than the actual label, in which case the outer circle is of a different colour than the colour of the inner point, and in the second case, the model prediction is same as the actual label, and the inner point and the outer circle have the same colours. Ideally, the colour of a particular point should match the background colour for the UMAP to be a correct representation of the decision boundaries. There are very few encircled points in all the figures of Fig. \ref{fig:decisionregion}, which indicates that UMAP is a good representation of the decision region.

When the samples are perturbed with adversarial noise, most of the samples are misclassified with baseline, and the decision regions are scattered, as seen from Fig. \ref{fig:baseline_adv}. Misclassified samples are those whose colours are different from the background colour in the case of non-encircled points and different from the outer circle colour for the encircled points. However, a significantly smaller number of adversarial points are misclassified while using ANF, as shown in Fig. \ref{fig:proposed_adv}. The corresponding plots with only clean samples are provided in subSec. D in the supplementary section. 

\subsection{Visualization of loss surfaces}
We look at the loss surface visualization utilizing the ideas in \cite{li2018visualizing}, which demonstrate that visualization is only possible in lower dimensions, such as using line or surface plots. After the training, the clean loss surface should take the form of a convex surface and have minima at the center of the plot $\mathbf{w}^*$ for both the baseline and the proposed architecture with ANF. In Fig. \ref{fig:loss_surface}, we study the loss surface when tested with adversarial samples. Fig. \ref{fig:losssurface_3d_baseline_adv} and Fig. \ref{fig:losssurface_3d_proposed_adv} represent the 3D adversarial loss surface of baseline and proposed ANF, respectively, from where it can be observed that the ANF has smoother loss surface and the convergence is better for ANF. To understand better, we present the contour plots of baseline and ANF in Fig. \ref{fig:losssurface_contour_baseline_adv} and Fig. \ref{fig:losssurface_contour_proposed_adv}, respectively. The contour plots depict that the loss with adversarial samples in the proposed model is $3.10$, which is much less when compared to the loss in the baseline model, which is $9.10$. Moreover, the contour plot of the adversarial loss for baseline demonstrates more than one minima, whereas the proposed ANF has one distinct minima. 

\subsection{Spectrum of adversarially perturbed features}
This study aims to elucidate the impact of ANF on noise and assess its potential as a filter for attenuating high-intensity frequency components. We observed that ANF effectively suppresses high-frequency components, unlike the baseline model. Due to space constraints, the spectrum with adversarial noise and corruption noise spectrum for ANF is discussed in detail in subSec. E of the supplementary section. 

Adversarial noise is considered structured noise because it is intentionally designed to exploit model vulnerabilities through targeted optimization, inducing specific errors. In contrast, unstructured noise, such as random or normal noise, lacks a specific pattern or intent, being randomly generated from a probability distribution without targeting model weaknesses. The authors in \cite{gilmer2019adversarial} claim that improving adversarial robustness against structured noise improves the robustness against image corruption, too. We observe that ANF does not only have better adversarial accuracy with respect to FGSM, PGD, and AA but is also resilient towards corruption noise, as highlighted in Table \ref{tab:resnet50}. Accuracies against corruption noise for various standard deviation $\sigma$ are also highlighted in subSec. E of the supplementary section.

\subsection{Feature denoising using ANF}
\cite{xie2019feature} suggest that the adversarial perturbations on images lead to noise in the features constructed by these networks. The authors have developed new network architectures that increase adversarial robustness by performing feature denoising. We show that ANF also removes the noises in the features, which we discuss in detail in subSec. F of the supplementary section.

\section{Conclusion}
We have shown that adversarial robustness can be achieved by designing a robust first layer, which acts as a first line of defense. We term the proposed first layer adversarial noise filter since it filters/suppresses adversarial noise implicitly. The ANF is composed of a combination of simple existing operations: a larger kernel, a higher number of filters, and a maxpool operation. We show that it leads to (a) higher adversarial accuracies for FGSM, PGD, and AA, (b) higher mPSNR, (c) better loss surfaces, d) robust decision boundaries, e) more robustness to corruption noise, etc., for a wide range of datasets and architectures. Note that all the SOTA architectures in native robustness typically change multiple layers, whereas, with only the first layer, we significantly outperform existing SOTA natively robust architectures.

\bibliographystyle{IEEEtran}
\bibliography{aaai25.bib}

\clearpage

\section{Supplementary}
In Sec. 4 of the main paper, we have discussed the effect of incorporating ANF as the first layer of different architectures, such as ResNet, WideResNet, VGG, and EfficientNet. We have shown architectural modifications only for ResNet 20 and WideResNet in the main paper due to space constraints, whereas the architectural changes in VGG16 due to introducing ANF are discussed here in detail. In all the experiments, ANF represents ANF with maxpool of stride one, i.e., ANF1, unless otherwise specified. Wherever we want to make a comparison across strides, we have used the terms ANF1 and ANF2 to represent ANF with maxpool of stride one and two, respectively.

\subsection{ANF with VGG16}
\begin{figure*}[h]
    \centering
    \includegraphics[scale=0.35]{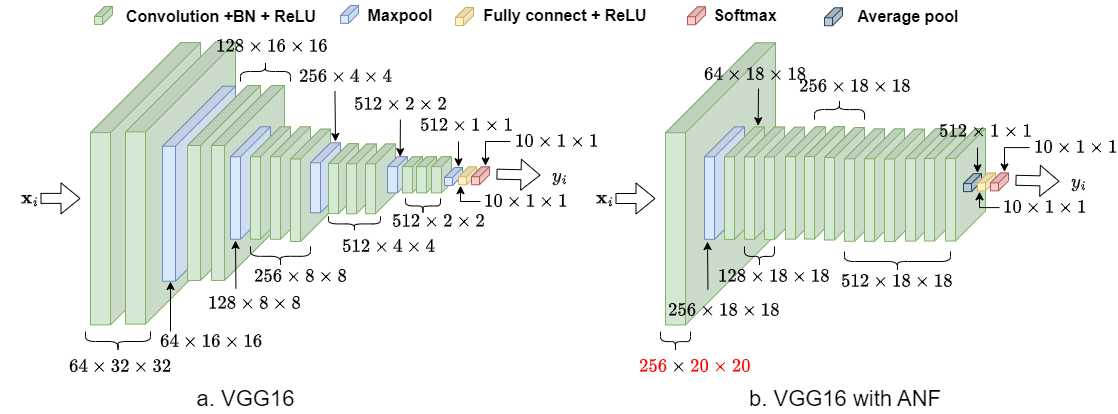}
    \caption{ANF implemented with VGG architecture, showcasing modified feature dimensions.}
    \label{fig:vgg}
\end{figure*}

In Sec. 4.2 of the main paper, we discuss how ANF is incorporated into VGG based on its structure and how ANF provides better adversarial robustness. In this section, we explain the modifications with the help of Fig. \ref{fig:vgg}. The VGG architecture on the left serves as the baseline, whereas the architecture on the right is with ANF, highlighting the proposed modifications, especially regarding feature dimensions. The baseline uses multiple maxpool layers to reduce the feature dimension across the depth of the network. The number corresponding to every layer shows the dimension of the output feature from that layer. As the VGG16 with ANF uses a bigger kernel size for convolution ($15$ instead of $3$ in the baseline), the dimension of the output feature from the convolution corresponding to ANF reduces to $20\times 20$ compared to $32\times 32$. Whereas the baseline uses multiple maxpool layers, we use only one maxpool in the architecture as part of the ANF.

In the proposed architecture, we use a stride of two for the max-pool layer corresponding to the ANF to achieve a similar reduction in feature dimension as the baseline. We call this version with a stride of two as ANF2 in Table \ref{tab:vgg_results}. We have also tested adversarial robustness with maxpool of stride one, called ANF1, in the same table. The accuracies of both the versions ANF1 and ANF2 for different attack methods are shown in Table. \ref{tab:vgg_results}. We observe that ANF2 outperforms ANF1, leading us to conclude that a higher stride in max pooling results in greater downsampling, which enhances the architecture's robustness against adversarial attacks.

\begin{table}[!ht]
    \centering
    \footnotesize 
    \setlength{\tabcolsep}{3mm} 
    \begin{tabular}{p{0.7cm}p{0.6cm}p{0.6cm}p{0.6cm}p{1.0cm}}
    \toprule
        \multicolumn{5}{c}{VGG16 with CIFAR10} \\
        \midrule
        Arch. & FGSM & PGD & AA  & Clean acc\\
        \midrule
        Baseline  & $60.60$ & $54.94$ & $43.83$ & $\textbf{92.42}$ \\
        ANF1 & $59.15$ & $60.51$ & $56.93$ & $80.66$ \\
        ANF2 & $\textbf{62.27}$ & $\textbf{63.11}$ & $\textbf{60.42}$ & $80.44$ \\
        
        \bottomrule
    \end{tabular}
    \caption{Comparison of different architectures on CIFAR10.}
    \label{tab:vgg_results}  
\end{table}

\subsection{Analysis of ANF hyperparameters across different adversarial attack intensities and corruption noise levels}
In this section, we change the ANF hyperparameters (K, F, and M) and observe the effect on adversarial robustness against multiple attacks such as FGSM, PGD, and AA for ResNet20 and CIFAR10. We always keep two hyperparameters fixed while varying the other hyperparameters. Among the hyperparameters in Table \ref{tab:variations_fgsm}, Table \ref{tab:variations_pgd}, and Table \ref{tab:variations_aa}, K, F, and M refer to convolutional kernel size, number of filters, and the kernel size of the maxpool operation. While in the main paper, we kept $\epsilon = 1/255$ similar to \cite{lukasik2023improving}, here we are varying $\epsilon$ from $1/255$ to $8/255$.

\subsubsection{FGSM}
Table \ref{tab:variations_fgsm} presents the impact of varying hyperparameters for ANF on the adversarial accuracy against Fast Gradient Sign Method (FGSM) attacks\footnote{https://github.com/bethgelab/foolbox} with multiple attack strengths. FGSM, a weaker adversarial attack than AA and PGD, shows different robustness across attack strengths. The table can be divided into three sub-parts to analyze these effects. From row 1 to row 3, it is observed that maxpool kernel size of $2\times 2$ provides greater robustness against stronger attacks like $4/255$ and $8/255$, while a maxpool kernel size of $3\times 3$ results in better robustness against lower attack strengths such as $1/255$ and $2/255$. In rows 4 to 6, the results focus on variations in K, showing that a convolution kernel size of $9\times 9$ offers superior adversarial robustness across almost all attack strengths. Rows 7 to 9  highlight a clear trend of enhanced adversarial accuracy across all attack strengths, with an increase in the number of filters F. 

\begin{table}[ht]
    \centering
    \begin{tabular}{p{0.4cm}p{1.0cm}p{0.7cm}p{0.6cm}p{0.6cm}p{0.6cm}p{0.6cm}p{0.5cm}}
    \toprule
        F & K & M &  $\frac{1}{255}$  & $\frac{2}{255}$  & $\frac{4}{255}$   & $\frac{8}{255}$ & Clean acc \\
    \midrule
     $256$ & $15\times15$& $2\times2$ & $61.57$ & $40.59$ & $\textbf{23.46}$ & $\textbf{14.10}$ & $\textbf{86.68}$  \\
     $256$ & $15\times15$ & $3\times3$ & $\textbf{61.76}$ & $\textbf{41.59}$ & $22.42$ & $12.72$ & $84.8$  \\
     $256$ & $15\times15$& $5\times5$ & $60.00$ & $40.18$ & $21.83$ & $12.35$ & $83.34$  \\
    \midrule
     $256$ & $5\times5$& $5\times5$ & $56.97$ & $36.18$ & $20.63$ & $11.90$ & $\textbf{89.63}$  \\
     $256$ & $9\times9$& $5\times5$ & $\textbf{60.67}$& $39.97$ & $\textbf{24.06}$ & $\textbf{14.24}$ & $87.39$  \\
     $256$ & $15\times15$& $5\times5$ & $60.00$ & $\textbf{40.18}$ & $21.83$ & $12.35$ & $83.34$  \\
    \midrule
     $64$ & $15\times15$& $5\times5$ & $55.73$ & $35.73$ & $17.87$ & $9.63$ & $82.25$  \\
     $128$ & $15\times15$& $5\times5$ & $57.88$ & $37.56$ & $18.80$ & $10.73$ & $82.58$  \\
     $256$ & $15\times15$& $5\times5$ & $\textbf{60.00}$ & $\textbf{40.18}$ & $\textbf{21.83}$ & $\textbf{12.35}$ & $\textbf{83.34}$\\
    \bottomrule
    \end{tabular}
    \caption{Accuracy as a function of  $\epsilon$, K, F, and M. The study is for ResNet20 on CIFAR10 classification for FGSM attack. The values indicate performance metrics, with bold values highlighting the best results.}
    \label{tab:variations_fgsm}
\end{table}

\subsubsection{Variations in PGD Attacks}
Table \ref{tab:variations_pgd} illustrates the effect of varying hyperparameters against the Projected Gradient Descent (PGD) attacks. The attacks are varied from $1/255$ to $8/255$ with $40$ iterations\footnote{https://github.com/bethgelab/foolbox}. From row 1 to row 3, it is observed that increasing the value of Maxpool kernel size from $2\times 2$ to $3\times 3$ enhances adversarial robustness, particularly at lower attack strengths such as $\epsilon = 1/255$, $2/255$, and $4/255$. This suggests that a larger M improves robustness against stronger attack strengths. Notably, when M is set to $5\times 5$, the robustness is significantly better at $\epsilon = 8/255$. This indicates that increased downsampling, due to a larger M, implicitly aids in resisting stronger attacks when the model is adequately trained. From row 4 to row 6, it is evident that increasing the convolutional kernel size K to $15\times 15$ consistently improves adversarial accuracy across all attack strengths. The margin of improvement becomes more pronounced as the attack strength increases, demonstrating the effectiveness of larger kernels in enhancing robustness. Row 7 to row 9 in Table \ref{tab:variations_pgd} follow a similar trend as observed with FGSM in Table \ref{tab:variations_fgsm}, where an increase in the number of filters, F, leads to a corresponding increase in adversarial accuracy.

\begin{table}[h]
    \centering
\begin{tabular}{p{0.4cm}p{1.0cm}p{1.0cm}p{0.5cm}p{0.5cm}p{0.5cm}p{0.5cm}p{0.5cm}}
    \toprule
    F & K & M &  $\frac{1}{255}$  & $\frac{2}{255}$  & $\frac{4}{255}$   & $\frac{8}{255}$ & Clean acc \\
    \midrule
     $256$ & $15\times15$& $2\times2$ & $61.09$ & $34.12$ & $8.25$ & $0.39$ & $\textbf{86.68}$  \\
     $256$ & $15\times15$& $3\times3$ & $\textbf{62.03}$ &  $\textbf{37.92}$ & $\textbf{11.62}$ & $0.80$ & $84.8$  \\
     $256$ & $15\times15$& $5\times5$ & $60.23$ & $36.79$ & $11.02$ & $\textbf{0.85}$ & $83.34$  \\
    \midrule
    $256$ & $5\times5$& $5\times5$ & $54.77$ & $23.13$ & $2.88$ & $0.0$ & $\textbf{89.63}$  \\
    $256$ & $9\times9$& $5\times5$ & $60.14$ & $32.06$ & $7.46$ & $0.32$ & $87.39$  \\
    $256$ & $15\times15$& $5\times5$ & $\textbf{60.23}$ & $\textbf{36.79}$ & $\textbf{11.02}$ & $\textbf{0.85}$ & $83.33$  \\
    \midrule
    $64$ & $15\times15$& $5\times5$ & $55.34$ & $30.59$ & $7.35$ & $0.39$ & $82.25$  \\
    $128$ & $15\times15$& $5\times5$ & $57.79$ & $33.19$ & $8.45$ & $0.49$ & $82.58$  \\
    $256$ & $15\times15$& $5\times5$ & $\textbf{60.23}$ & $\textbf{36.79}$ & $\textbf{11.02}$ & $\textbf{0.85}$ & $\textbf{83.33}$  \\
    \bottomrule
    \end{tabular}
    \caption{Accuracy as a function of  $\epsilon$, K, F, and M. The study is for ResNet20 on CIFAR10 classification for PGD attack. The values indicate performance metrics, with bold values highlighting the best results}
    \label{tab:variations_pgd}
\end{table}

\begin{table*}[!ht] 
    \centering
    \begin{tabular}{p{1.0cm}p{0.9cm}p{0.8cm}p{0.8cm}p{0.9cm}p{0.8cm}p{0.8cm}p{0.9cm}p{0.8cm}p{0.8cm}}
        \toprule
        \multirow{1}*{$\epsilon$ } & \multicolumn{3}{c}{PGD} & \multicolumn{3}{c}{FGSM}  & \multicolumn{3}{c}{AA}  \\
        \cmidrule(lr){2-4} \cmidrule(lr){5-7} \cmidrule(lr){8-10} 
         & Baseline & ANF1 & ANF2 & Baseline & ANF1 & ANF2 & Baseline & ANF1 & ANF2 \\
         \midrule

         $1/255$ & $26.26$ & $60.23$ & $\textbf{61.21}$ & $43.10$ & $60.00$ & $\textbf{61.17}$ & $11.40$ & $55.23$ & $\textbf{56.56}$ \\
         $2/255$ & $3.55$ & $36.79$ & $\textbf{39.17}$ & $29.00$ & $40.18$ & $\textbf{43.21}$ & $0.06$ & $27.12$ & $\textbf{29.06}$ \\
         $4/255$ & $0.02$ & $11.02$ & $\textbf{11.80}$ &  $20.72$ & $\textbf{21.83}$ & $21.66$ & $0.00$ & $2.97$ & $\textbf{3.57}$ \\
         
         \bottomrule
    \end{tabular}
    \caption{Adversarial accuracies for ResNet20 with variation in $\epsilon$ for adversarial noise for variation in strides.}
    \label{tab:variation_stride}
\end{table*}

\subsubsection{Variations in AA Attacks}
Table \ref{tab:variations_aa} presents the impact of varying ANF hyperparameters against Auto Attacks (AA) \footnote{https://github.com/fra31/auto-attack.git} with $\epsilon$ varied from $1/255$ to $8/255$. From rows 1 to 3, the results indicate that increasing the kernel size M to $5\times5$ yields promising results for $\epsilon = 1/255$. However, a kernel size of $3\times3$ for higher attack strengths demonstrates slightly better performance. It is important to note that the difference in accuracy between the $3\times3$ and $5\times5$ configurations is minimal in the case of AA, especially when compared to the observations made in the PGD and FGSM studies. Row 4 to row 6 mirrors the trend observed in PGD, as shown in Table \ref{tab:variations_pgd}. Specifically, increasing the convolutional kernel size K to $15\times15$ consistently enhances adversarial robustness across all attack strengths, outperforming other configurations. In row 7 to row 9, a similar pattern is observed with FGSM and PGD: increasing the number of filters, F improves adversarial accuracy. This consistent trend underscores the importance of filter size in bolstering model robustness against AA.

\begin{table}[!t]
    \centering
    \begin{tabular}{p{0.8cm}p{1.0cm}p{0.8cm}p{0.5cm}p{0.5cm}p{0.5cm}p{0.5cm}p{0.5cm}}
    \toprule
       F & K & M &   $\frac{1}{255}$  & $\frac{2}{255}$  & $\frac{4}{255}$   & $\frac{8}{255}$ & Clean acc \\
    \midrule
    $256$ & $15\times15$& $2\times2$ & $55.1$ & $22.53$ & $1.44$ & $0.01$ & $\textbf{86.68}$  \\
    $256$ & $15\times15$& $3\times3$ & $55.11$ & $\textbf{28.14}$ & $\textbf{2.99}$ & $\textbf{0.03}$ & $84.8$  \\
    $256$ & $15\times15$& $5\times5$ & $\textbf{55.23}$ & $27.12$ & $2.97$ & $0.00$ & $83.34$  \\
    \midrule
     $256$ & $5\times5$& $5\times5$ & $45.89$ & $10.49$ & $0.11$ & $0.01$ & $\textbf{89.63}$  \\
    $256$ & $9\times9$& $5\times5$ & $53.7$ & $19.66$ & $0.95$ & $\textbf{0.02}$ & $87.39$  \\
    $256$ & $15\times15$& $5\times5$ & $\textbf{55.23}$ & $\textbf{27.12}$ & $\textbf{2.97}$ & $0.00$ & $83.34$  \\
    \midrule
    $64$ & $15\times15$& $5\times5$ & $49.55$ & $20.72$ & $1.68$ & $0.00$ & $82.25$  \\
    $128$ & $15\times15$& $5\times5$ & $52.55$ & $22.94$ & $1.79$ & $0.00$ & $82.58$  \\
    $256$ & $15\times15$& $5\times5$ & $\textbf{55.23}$ & $\textbf{27.12}$ & $\textbf{2.97}$ & $0.00$ & $\textbf{83.34}$  \\
    \bottomrule
    \end{tabular}
    \caption{Accuracy as a function of  $\epsilon$, K, F, and M. The study is for ResNet20 on CIFAR10 classification. The values indicate performance metrics, with bold values highlighting the best results}
    \label{tab:variations_aa}
\end{table}

\begin{figure*}[t]
    \centering
    \begin{subfigure}{0.43\textwidth}
    \includegraphics[width=\textwidth]{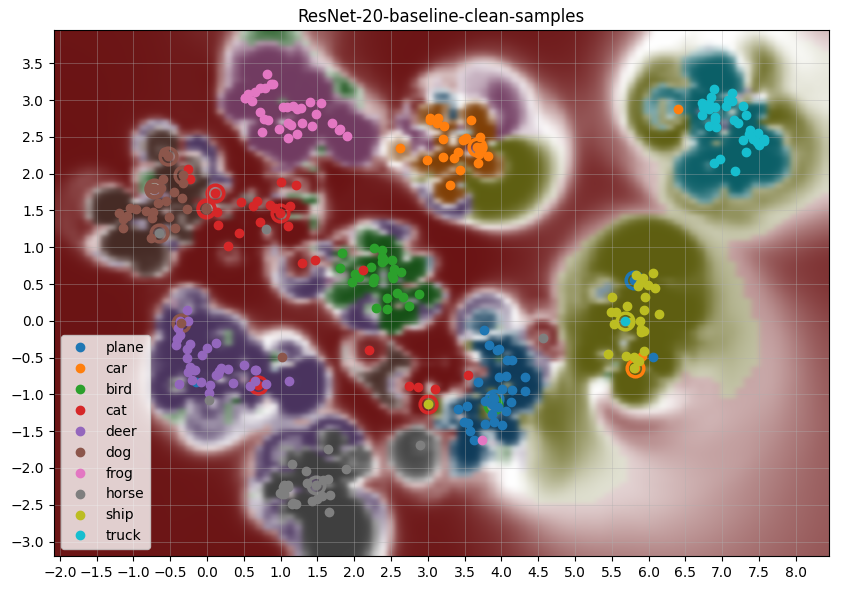}
    \caption{Baseline with clean samples}
    \label{fig:baseline_clean}
    \end{subfigure}%
    \begin{subfigure}{0.40\textwidth}
    \includegraphics[width=\textwidth]{fig/baseline_adv_samples_desc_boundary.png}
    \caption{Baseline with adversarial samples}
    \label{fig:baseline_adv_supp}
    \end{subfigure}
    \begin{subfigure}{0.41\textwidth}
    \includegraphics[width=\textwidth]{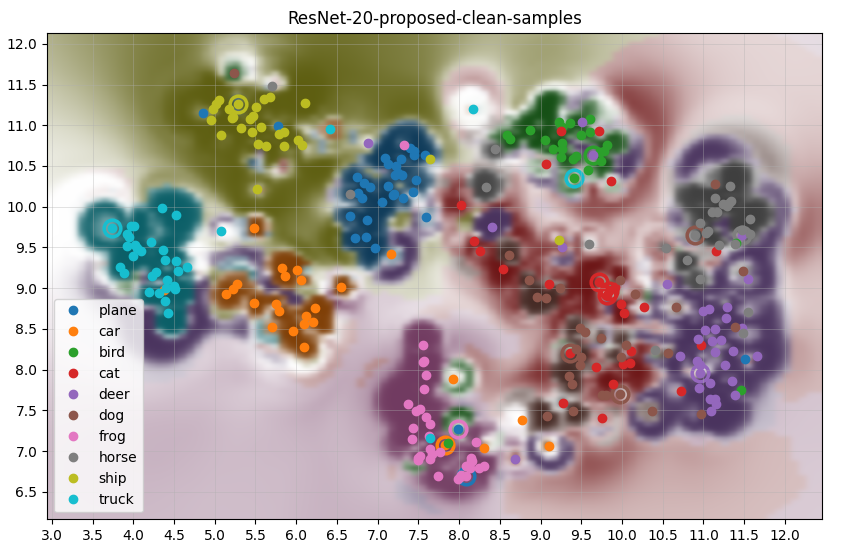}
    \caption{ANF with clean samples}
    \label{fig:proposed_clean}
    \end{subfigure}%
    \begin{subfigure}{0.45\textwidth}
    \includegraphics[width=\textwidth]{fig/proposed_adv_samples_desc_boundary.png}
    \caption{ANF with adversarial samples}
    \label{fig:proposed_adv_supp}
    \end{subfigure} 
    \caption{Visualization of the decision regions}
    \label{fig:decisionregion_supp}
\end{figure*}

\subsubsection{Variations in Unstructured Noise or Corruption Noise}

Table \ref{tab:variations_unstructure_noise} presents the impact of varying hyperparameters on model performance under corruption noise with different $\sigma$ values. Note that $\sigma$ is the standard deviation of the Gaussian noise. Unlike adversarial noise, corruption noise exhibits distinct behavior in response to hyperparameter variations. Row 1 to row 3 shows that the highest corruption accuracy is achieved with a kernel size M of $2\times2$. As M increases, there is a noticeable decline in accuracy, particularly at higher values. Row 4 to row 6 demonstrates that the optimal kernel size K for most $\sigma$ variations is $5\times5$. However, for $\sigma = 16/255$, the best accuracy is observed with K set to $9\times9$. For higher noise strengths, the performance of K at $9\times9$ closely approximates that of $15\times15$. Rows 7 to 9 reveal a trend consistent with the observations made for other adversarial attacks: increasing the number of filters, F, generally improves corruption accuracy. This suggests that the benefits of increasing F extend beyond adversarial robustness also to enhance performance under corruption noise.

\begin{table}[h]
    \centering
    \begin{tabular}{p{0.5cm}p{1.0cm}p{0.8cm}p{0.5cm}p{0.6cm}p{0.5cm}p{0.5cm}p{0.5cm}}
    \toprule
        F & K & M&  $\frac{1}{255}$  & $\frac{2}{255}$  & $\frac{4}{255}$   & $\frac{8}{255}$ & $\frac{16}{255}$ \\
    \midrule
    $256$ & $15\times15$& $2\times2$ & $\textbf{85.19}$ & $\textbf{85.18}$ & $\textbf{85.15}$ & $\textbf{84.71}$ & $\textbf{81.64}$  \\
    $256$ & $15\times15$& $3\times3$ & $83.07$ & $83.27$ & $82.78$ & $82.54$ & $80.05$  \\
    $256$ & $15\times15$& $5\times5$ & $81.8$ & $81.92$ & $81.64$ & $81.45$ & $79.51$  \\
    \midrule
    $256$ & $5\times5$& $5\times5$ & $\textbf{86.79}$ & $\textbf{86.69}$ & $\textbf{86.6}$ & $\textbf{85.2}$ & $78.2$  \\
    $256$ & $9\times9$& $5\times5$ & $85.8$ & $85.79$ & $85.42$ & $84.3$ & $\textbf{80.81}$  \\
    $256$ & $15\times15$& $5\times5$ & $81.8$ & $81.92$ & $81.64$ & $81.45$ & $79.51$  \\
    \midrule
    $64$ & $15\times15$& $5\times5$ & $80.47$ & $80.44$ & $80.27$ & $79.91$ & $77.26$  \\
    $128$ & $15\times15$& $5\times5$ & $80.99$ & $80.87$ & $80.64$ & $80.04$ & $78.21$  \\
    $256$ & $15\times15$& $5\times5$ & $\textbf{81.8}$ & $\textbf{81.92}$ & $\textbf{81.64}$ & $\textbf{81.45}$ & $\textbf{79.51}$  \\
    \bottomrule
    \end{tabular}
    \caption{Accuracy as a function of  $\sigma$, K, F, and M. The study is for ResNet20 on CIFAR10 classification for corruption noise. The values indicate performance metrics, with bold values highlighting the best results}
    \label{tab:variations_unstructure_noise}
\end{table}

In summary, increasing the kernel size K to $15\times15$ is highly effective against stronger attacks such as PGD and AA and against unstructured noise with higher $\sigma$ values. Adjusting M to $3\times3$ enhances robustness for most adversarial attacks, while further increasing M to $5\times5$ can be particularly beneficial under stronger attack scenarios, such as with AA or with a higher $\epsilon$ for other attacks. Additionally, we consistently observe that increasing the number of filters, combined with larger K and M, improves accuracy against corruption noise and adversarial attacks.

\subsection{Adversarial robustness with varying maxpool stride}
In this section, we vary ANF's maxpool stride to see how much it impacts the adversarial accuracy, as shown in Table \ref{tab:variation_stride}. ANF1 and ANF2 are the variations of ANF with stride $1$ and $2$, respectively. We have taken ResNet20 architecture for this experiment with CIFAR10 dataset. We observe better adversarial accuracies with ANF2 for stronger attacks like PGD and AA for all the attack strengths. For weaker attacks like FGSM, ANF2 is much better at weaker strength, whereas when the attack strength increases, accuracy for ANF1 and Baseline improves. For corruption noise, particularly at $\sigma=16/255$ (where $\sigma$ represents the standard deviation), ANF1 achieves a corruption accuracy of $79.51$, the Baseline achieves $72.54$, and ANF2 achieves $80.15$, indicating that ANF2 is more effective in handling corruption noise at higher $\sigma$ values. To summarize, ANF2 provides a higher degree of downsampling, leading to better adversarial accuracies, especially for stronger attacks like PGD and AA. We have also reported the clean accuracies for Baseline, ANF1, and ANF2 in Table. \ref{tab:stride_clean} where we observe that ANF1 clean accuracy is marginally better than ANF2 because of a coarser downsampling induced by ANF2.

 
\begin{table}[h]
    \centering
    \begin{tabular}{cccc}
        \toprule
        Architecture & Baseline & ANF1 & ANF2 \\
        \midrule
        ResNet20 & $91.21$ & $83.34$ & $83.23$ \\
        \bottomrule
    \end{tabular}
    \caption{Clean accuracy for baseline, ANF1 and ANF2}
    \label{tab:stride_clean}
\end{table}

\subsection{Descision boundaries}
In Section 5.1 of the main paper, we examined the decision surfaces of the baseline model and the proposed ANF model in the context of adversarial samples, as illustrated in Fig. \ref{fig:baseline_adv_supp} and Fig. \ref{fig:proposed_adv_supp} respectively. In this section, we will discuss the decision surfaces of both the baseline and ANF architectures when evaluated on clean samples.
Figures \ref{fig:baseline_clean} and \ref{fig:proposed_clean} illustrate the decision regions of the baseline and proposed ANF architectures, respectively, when applied to clean samples from the CIFAR10 dataset using ResNet20. In these figures, the true labels, represented by filled circles, are overlaid on a background where the colors indicate the model's predictions. The decision regions in Figure \ref{fig:baseline_clean} are more distinctly defined, as demonstrated by the higher alignment between the true labels and the predictions of the baseline model. Conversely, Figure \ref{fig:proposed_clean} exhibits a reduced alignment between the true labels and the predictions for the proposed ANF model.

As noted by \cite{mcinnes2018umap}, the intensity of the background colors corresponds to the model's confidence, with darker colors signifying higher confidence in predictions. In summary, the analysis indicates that the baseline model's clean accuracy is anticipated to exceed that of the proposed ANF model. This expectation is confirmed by empirical results, with the baseline model achieving a clean accuracy of $91.26\%$, compared to $83.09\%$ for the ANF incorporated model.
\subsection{Adversarial and corruption noise frequency spectrum of ANF }
In Sec. 5.3 of the main paper, we investigate the spectrum of the perturbed features with adversarial and corruption noise. The spectrum is plotted so that the low-frequency components are towards the center of the spectrum and high-frequency components are towards the edge for better visualization \cite{lukasik2023improving}. 

\begin{figure*}[!ht]
    \centering
        \begin{subfigure}{0.25\textwidth}
        \includegraphics[width=\textwidth]{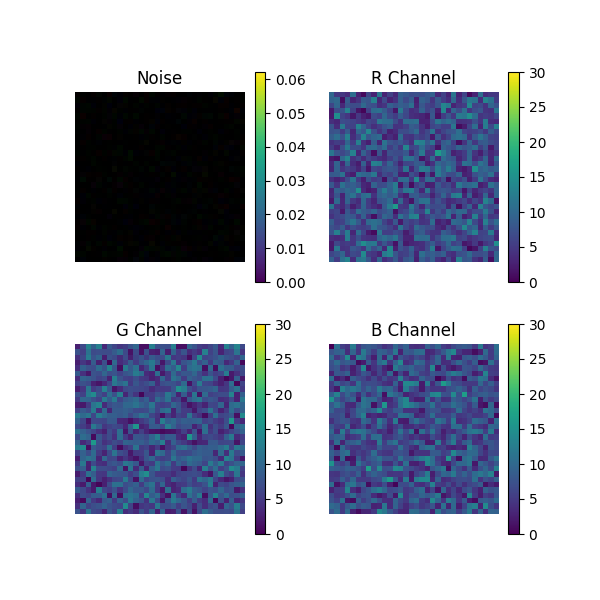}
        \caption{Baseline, At the input}
        \label{fig:baseline_corr_pre}
        \end{subfigure}
        \begin{subfigure}{0.5\textwidth}
        \includegraphics[width=\textwidth]{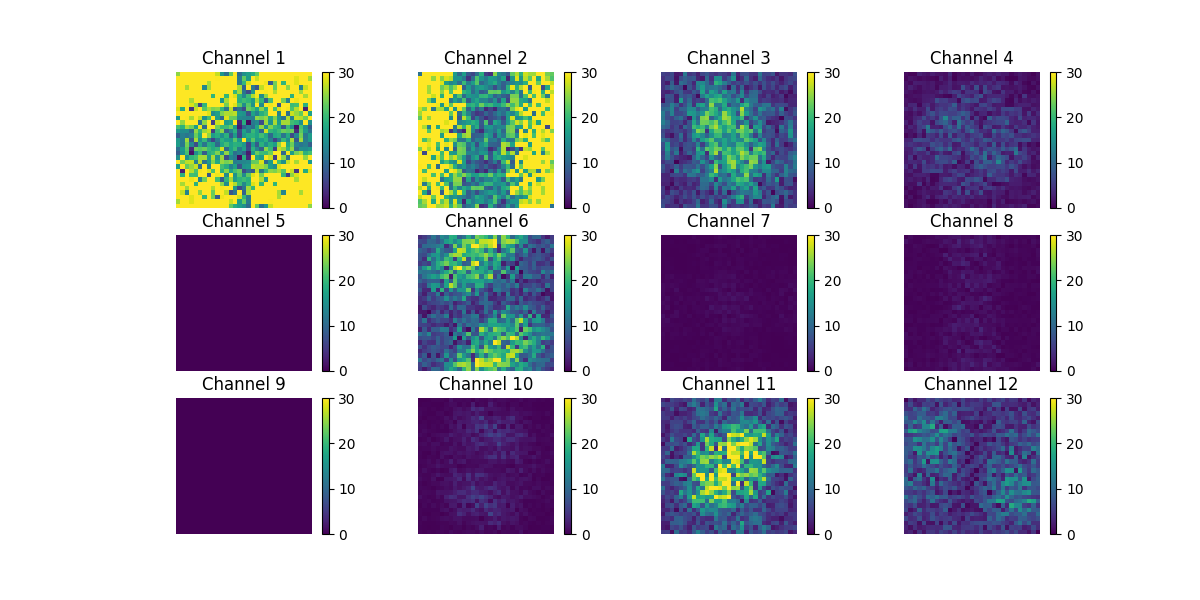}
        \caption{Baseline, After the first layer}
        \label{fig:baseline_corr_post}
        \end{subfigure}
        \begin{subfigure}{0.25\textwidth}
        \includegraphics[width=\textwidth]{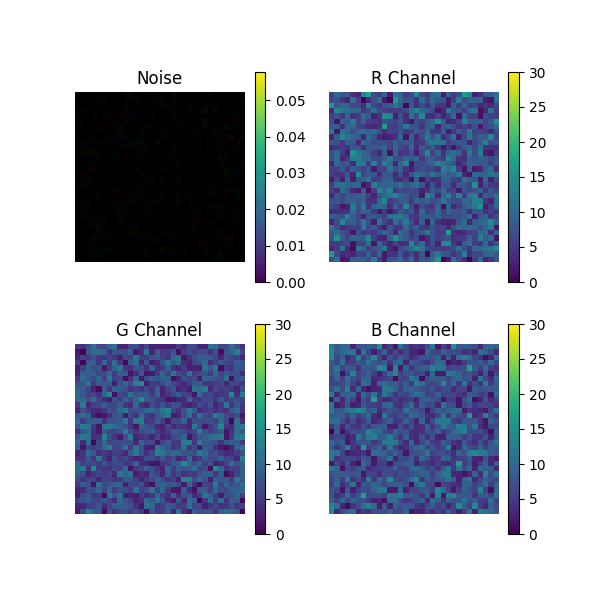}
        \caption{ANF, At the input}
        \label{fig:proposed_corr_pre}
        \end{subfigure}
        \begin{subfigure}{0.50\textwidth}
        \includegraphics[width=\textwidth]{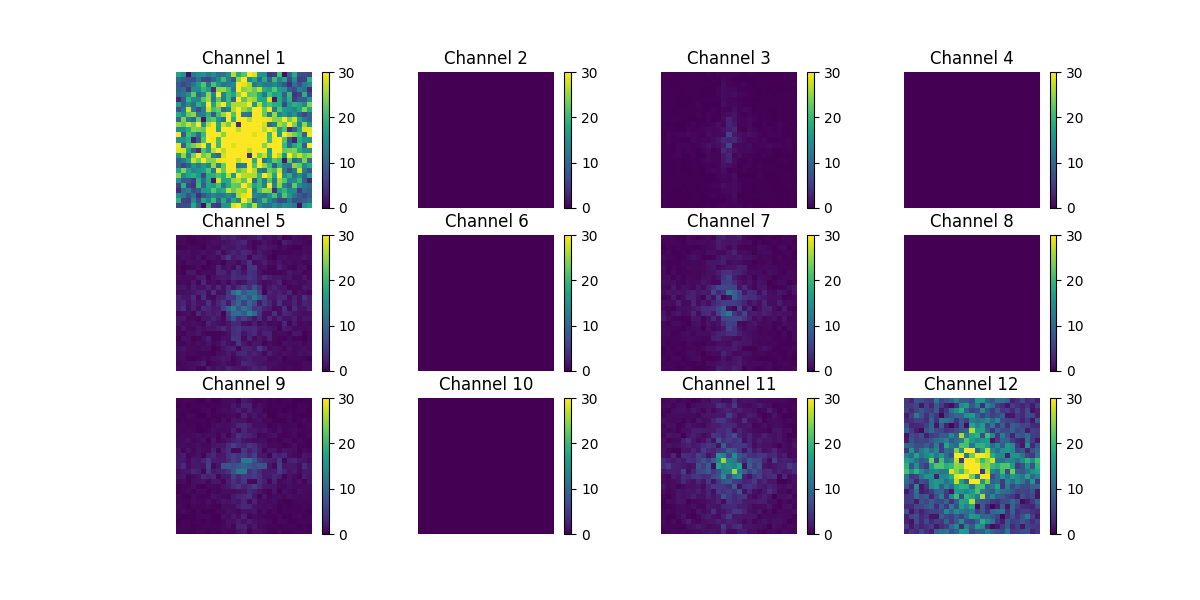}
        \caption{ANF, After the first layer}
        \label{fig:proposed_corr_post}
        \end{subfigure}%
    \caption{Spectrum of corruption noise at the input and after the first layer. The top row is for Baseline, and the bottom row is for ANF.}
    \label{fig:corruptionnoise}
\end{figure*}

The input with corruption noise does not have any pattern in the spectrum for both baseline and ANF, as shown in Fig. \ref{fig:baseline_corr_pre} and in Fig. \ref{fig:proposed_corr_pre}, respectively. However, the output noise spectrum after the first layer of baseline has some pattern with high magnitude in the high frequencies. Fig. \ref{fig:baseline_corr_post} and Fig. \ref{fig:proposed_corr_post} show the spectrum in the first $12$ channels of baseline and ANF, respectively, for corruption noise. Compared to the baseline architecture, the noise output after the ANF has a pattern that shows high intensity only for very few low-frequency components and very low intensity for most high-frequency components. We also show in table \ref{tab:unstructurednoise} how the corruption noise varies with different $\sigma$ for the baseline and the proposed EfficientNet-B0 architecture. Our ANF is a strong defense for higher $\sigma$, such as, $16/255$.

\begin{figure*}[!ht]
    \centering
        \begin{subfigure}{0.25\textwidth}
        \includegraphics[width=\textwidth]{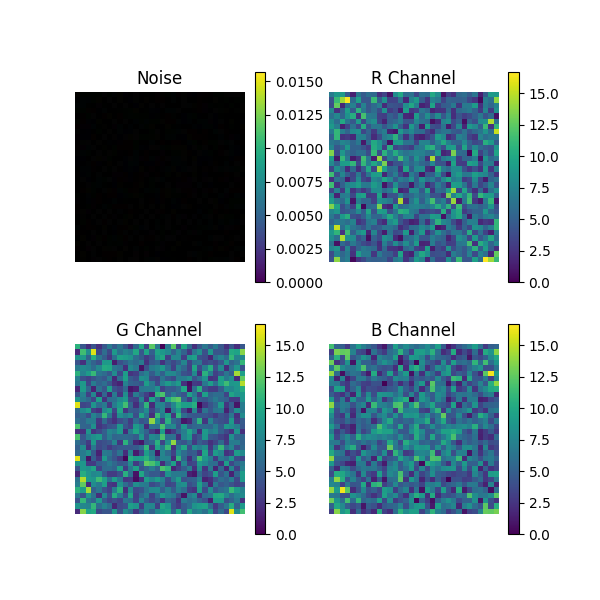}
        \caption{Baseline, At the input}
        \label{fig:baseline_adv_pre}
        \end{subfigure}%
        \begin{subfigure}{0.50\textwidth}
        \includegraphics[width=\textwidth]{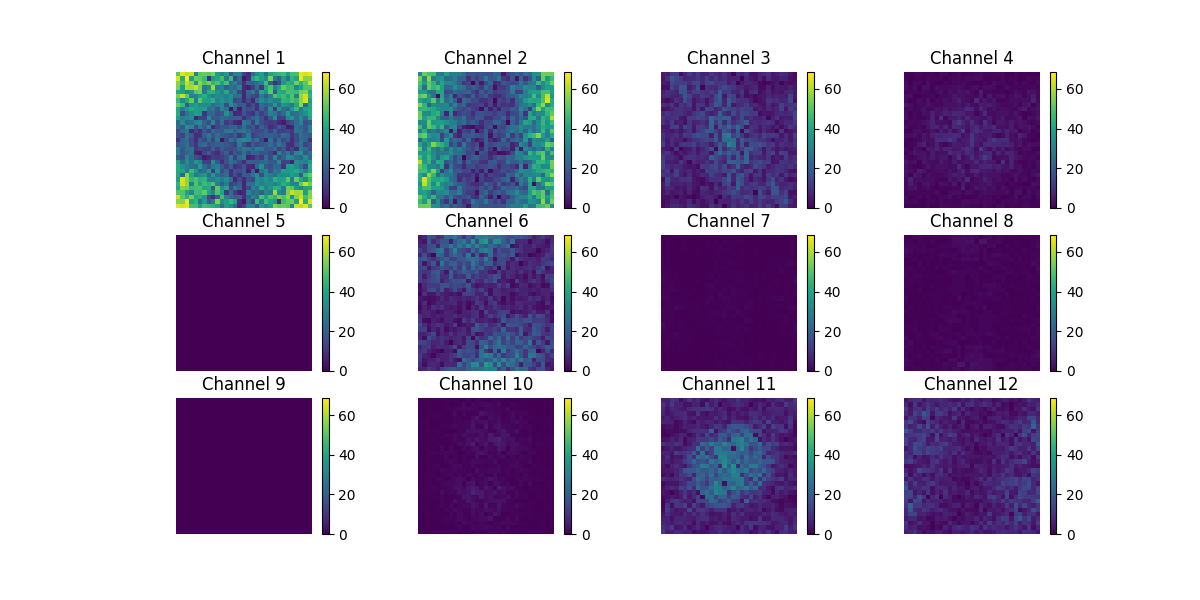}
        \caption{Baseline, After the first layer}
        \label{fig:baseline_adv_post}
        \end{subfigure}
        \begin{subfigure}{0.25\textwidth}
        \includegraphics[width=\textwidth]{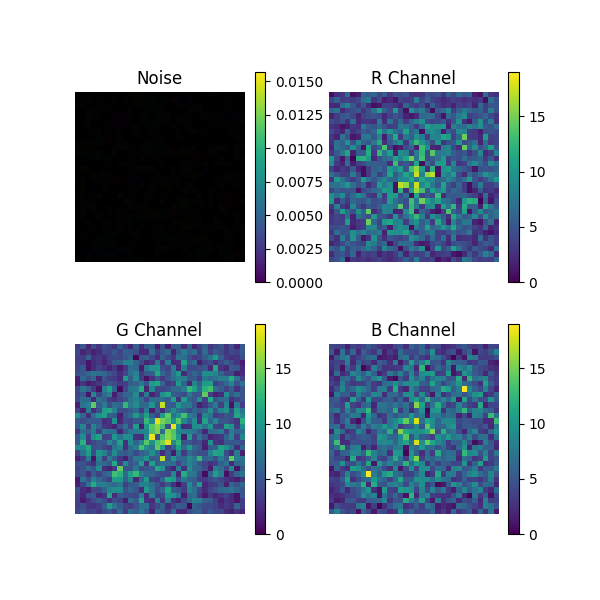}
        \caption{ANF, At the input}
        \label{fig:proposed_adv_pre}
        \end{subfigure}%
        \begin{subfigure}{0.50\textwidth}
        \includegraphics[width=\textwidth]{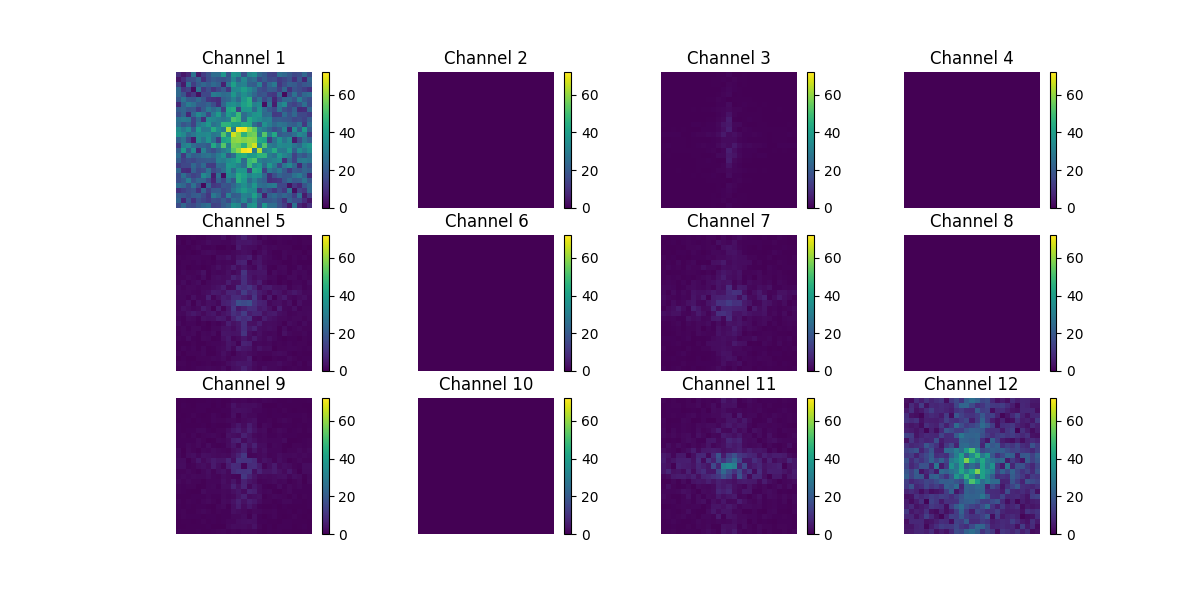}
        \caption{ANF, After the first layer}
        \label{fig:proposed_adv_post}
        \end{subfigure}%
    \caption{Spectrum of adversarial noise at the input and after the first layer.}
    \label{fig:spectrum_adversarial_noise}
\end{figure*}

The adversarial noise and its spectrum in both R, G, and B channels for baseline are shown in Fig. \ref{fig:baseline_adv_pre} and the same for ANF is shown in Fig. \ref{fig:proposed_adv_pre}. The structured adversarial noise is obtained by perturbing the input with PGD attack with $\epsilon=4/255$ and $40$ iterations. Fig. \ref{fig:baseline_adv_post} and Fig. \ref{fig:proposed_adv_post} show the spectrum in the first $12$ channels of baseline and ANF, respectively, for adversarial noise. The spectrum exhibits high-intensity components in the baseline model's input layer at high frequencies. In contrast, the input layer of the architecture incorporating ANF shows high intensity only at very low frequencies. When examining the propagation of noise through the ANF, one can observe that after the first layer, even though the baseline has high-frequency components with higher magnitudes, after using ANF, the spectrum has only very low-frequency components, and most of the high-frequency components are attenuated. Note that the noise for the baseline and the ANF are the same strength.

\begin{table}[!ht]
    \centering
    \footnotesize 
    \setlength{\tabcolsep}{3mm} 
    \begin{tabular}{p{0.6cm}p{0.4cm}p{0.4cm}p{0.4cm}p{0.4cm}p{0.4cm}p{0.4cm}p{0.4cm}}
        \toprule
        \multirow{2}*{Arch.} & \multicolumn{7}{c}{p} \\
        \cmidrule(lr){2-8}
        & $0.25$ & $0.5$ & $1$ & $2$ & $4$ & $8$ & $16$ \\
        \midrule
        Baseline & $92.24$ & $92.17$ & $92.11$ & $91.85$ & $90.14$ & $80.48$ & $44.08$ \\
        ANF & $\mathbf{87.16}$ & $\mathbf{87.16}$ & $\mathbf{87.18}$ & $\mathbf{87.08}$ & $\mathbf{87.01}$ & $\mathbf{86.11}$ & $\mathbf{80.18}$ \\
        \bottomrule
    \end{tabular}
    \caption{Accuracy (in \%) with unstructured noise where baseline is EfficientNet-B0. Standard deviation of the corruption noise is $\sigma=p/255$.}
    \label{tab:unstructurednoise}
\end{table}

\begin{figure}[!ht]
    \centering
    \begin{subfigure}{0.5\textwidth}
    \includegraphics[width=\textwidth]{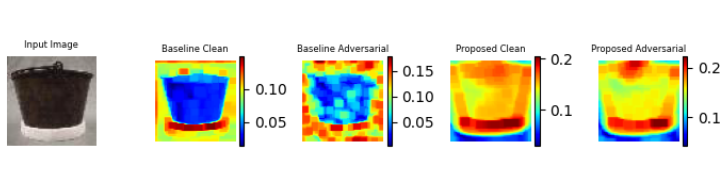}
    \vspace{-10mm}
    \caption{Image 1}
    \label{fig:featuremap_1}
    \end{subfigure}
    
    \begin{subfigure}{0.5\textwidth}
    \includegraphics[width=\textwidth]{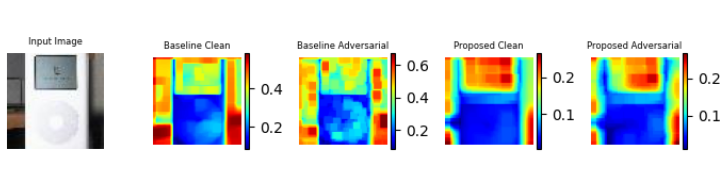}
    \vspace{-10mm}
    \caption{Image 2}
    \label{fig:featuremap_2}
    \end{subfigure}
    \caption{Feature maps of ResNet50 with TinyImagenet}
    \label{fig:featuremaps2}
\end{figure}

\subsection{Feature maps}
In this experiment, we investigate the appearance of feature maps for selected images. According to \cite{xie2019feature}, filters that smooth feature maps can also mitigate adversarial noise within these maps, thereby enhancing adversarial robustness. Figures \ref{fig:featuremap_1} and \ref{fig:featuremap_2} present two such feature maps.\footnote{Due to the differing number of channels in ANF, the feature maps obtained for ANF differ from those of the baseline model. We have selected feature maps that most effectively illustrate our observations.} In Figure \ref{fig:featuremaps2}, which displays feature maps for two distinct images, Figure \ref{fig:featuremap_1} shows that the feature map of the baseline model under an adversarial attack is considerably noisier compared to its clean counterpart. Conversely, the ANF feature map under an adversarial attack has filtered out most of the noise, closely resembling the clean feature map of ANF. A similar trend is observed in Figure \ref{fig:featuremap_2}, where the feature map of the adversarial sample exhibits significant noise, particularly in the blue regions, compared to the clean feature map of the baseline model. In contrast, the feature map of our proposed ANF model remains very close to its clean counterpart.

Because one of the components of ANF is maxpool, ANF can smooth out the noise by downsampling, which the baseline cannot do much. However, because of the downsampling operation, the details of the input feature get lost. Therefore, increasing the filters helps retain the input feature's details. 

\begin{figure*}[!ht]
\centering
\begin{subfigure}{0.5\textwidth}
    \includegraphics[width=\textwidth]{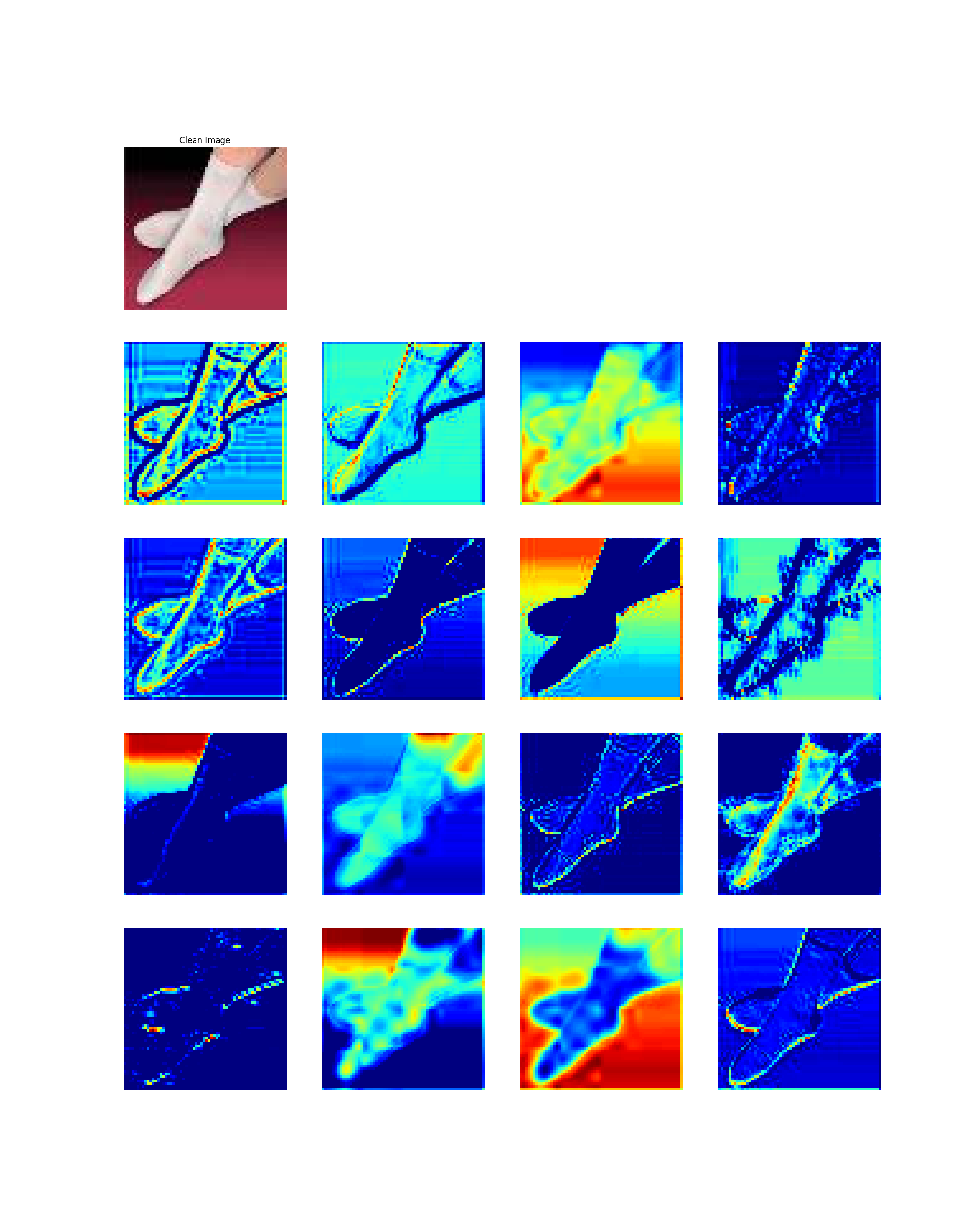}
    \caption{Baseline feature maps for first 16 filters for clean samples}
    \label{fig:baselinefeaturemapsclean}
\end{subfigure}%
\begin{subfigure}{0.5\textwidth}
    \includegraphics[width=\textwidth]{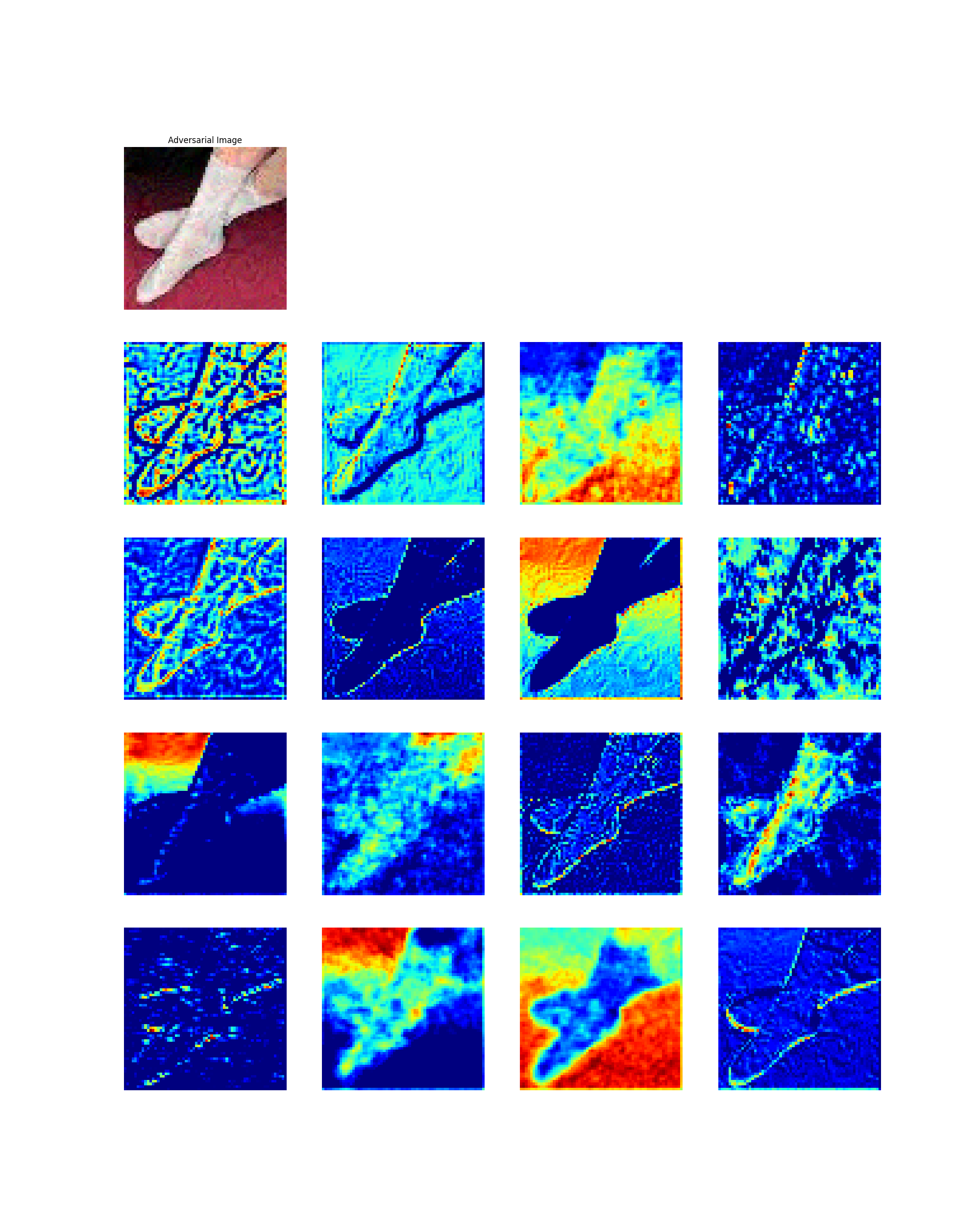}
    \caption{Baseline feature maps for first 16 filters for adversarial samples}
    \label{fig:baselinefeaturemapsadv}
\end{subfigure}
\begin{subfigure}{0.5\textwidth}
    \includegraphics[width=\textwidth]{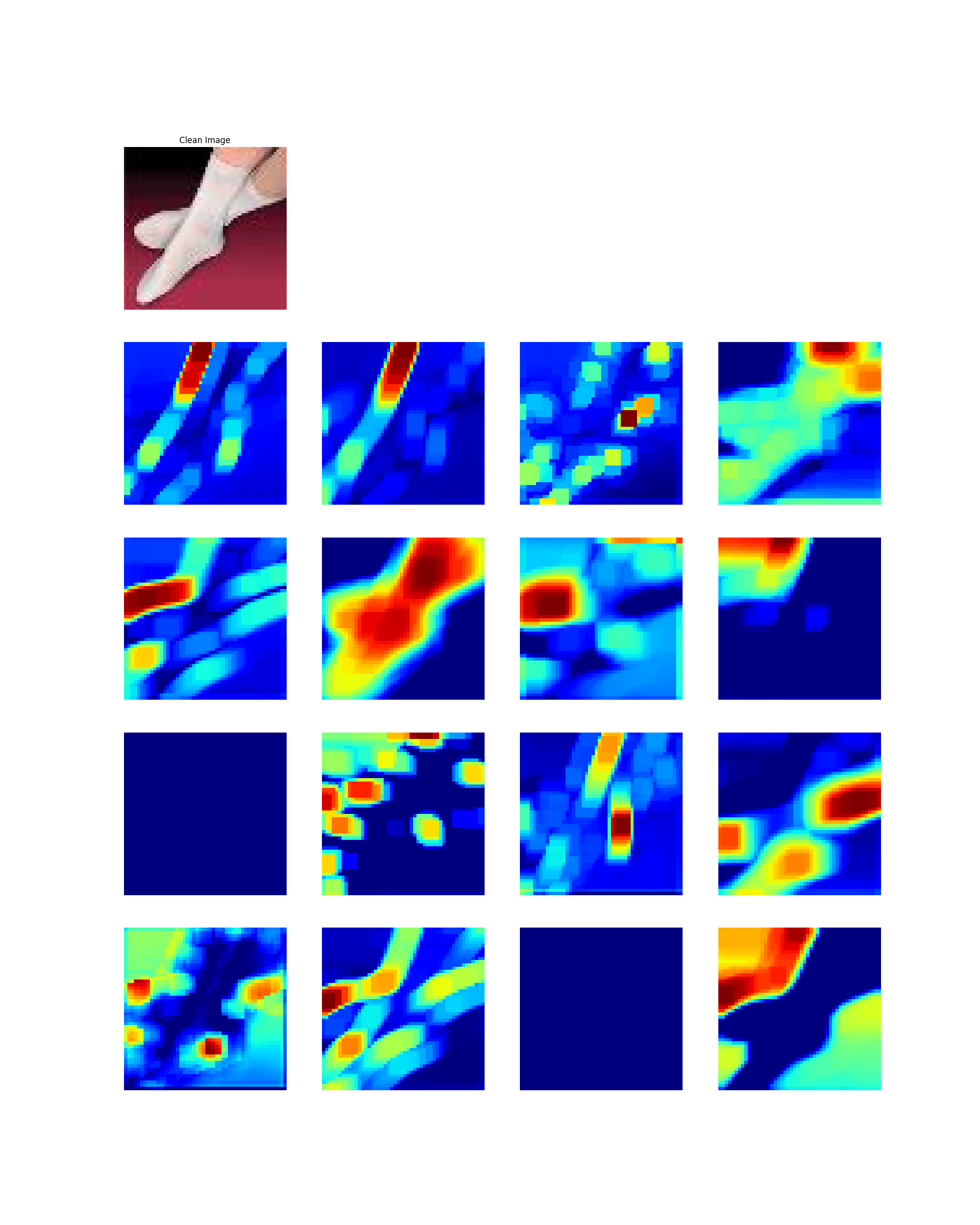}
    \caption{Proposed feature maps for first 16 filters for clean samples}
    \label{fig:proposedfeaturemapsclean}
\end{subfigure}%
\begin{subfigure}{0.5\textwidth}
    \includegraphics[width=\textwidth]{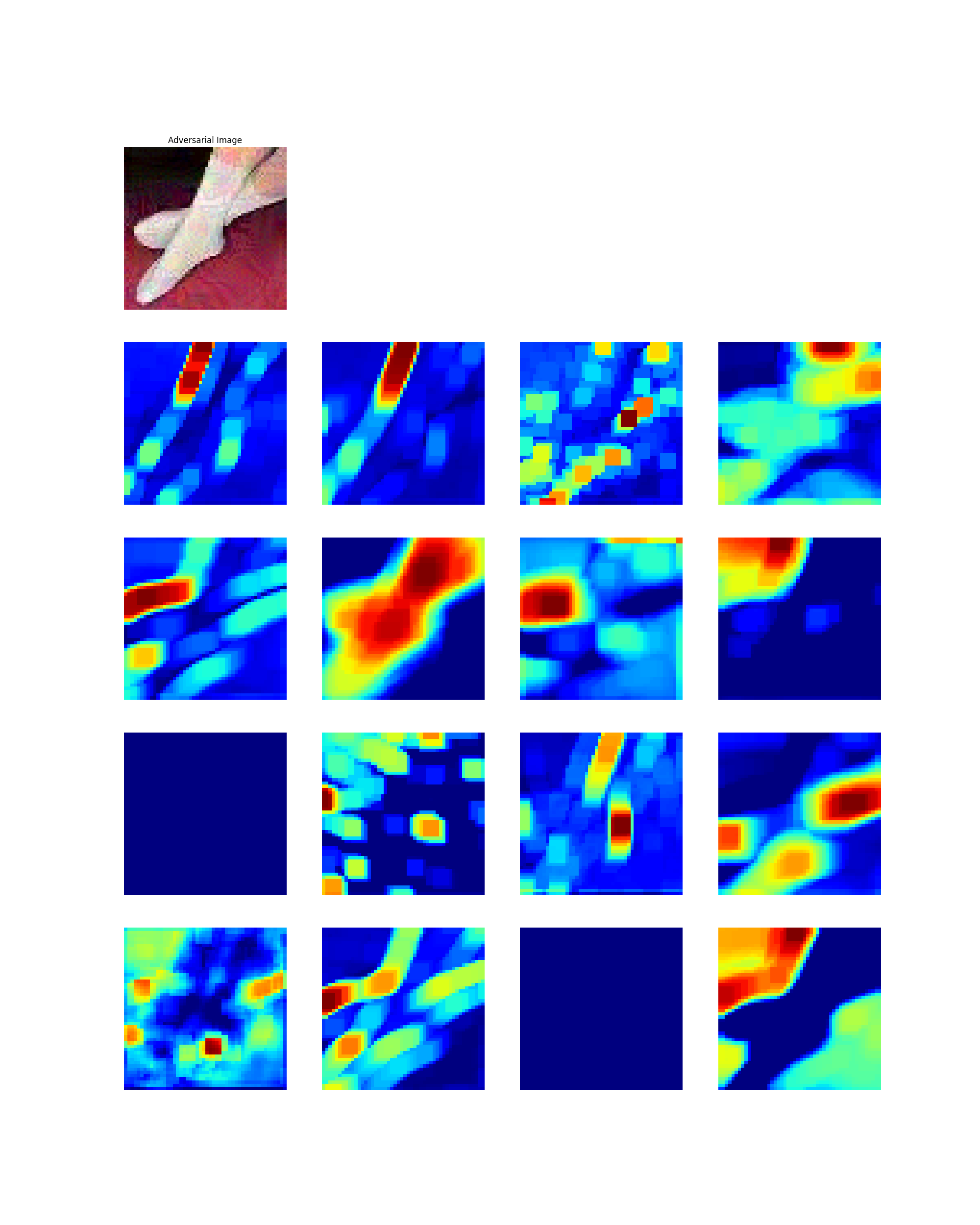}
    \caption{Proposed feature maps for first 16 filters for adversarial samples}
    \label{fig:proposedfeaturemapsadv}

\end{subfigure}
\caption{Visualisation of feature maps}
\end{figure*}

We also visualize the feature maps for the first 16 filters of ResNet50 for both clean and adversarial images for a different image. Figures \ref{fig:baselinefeaturemapsclean} and \ref{fig:baselinefeaturemapsadv} show the feature maps for the baseline architecture with clean and adversarial images, respectively. In contrast, Figures \ref{fig:proposedfeaturemapsclean} and \ref{fig:proposedfeaturemapsadv}display the feature maps for the proposed architecture under the same conditions. The feature maps produced by the proposed architecture exhibit minimal differences between clean and adversarial images, as our ANF effectively filters out the noise, resulting in smoother maps. On the other hand, the baseline architecture's feature maps for adversarial images appear significantly noisier than those for clean images. 

The adversarial noise used in this experiment is $\epsilon$ $16/255$ to make the noise visible to the naked eye.

\subsection{Attacks and Variations of other hyperparameters}
\subsubsection{Hyperparameters}
\label{section:hyperparameters}
Overall, the simulations are carried out for four different datasets. CIFAR10, CIFAR100, TinyImagenet and Imagenet. For CIFAR10 and CIFAR100, Imagenet we ran the simulations for 200 epochs, while TinyImagenet was run with 300 epochs. 

For EfficientNet-B0, the Adam optimizer was employed, utilizing a weight decay of 1e-2 and cosine annealing scheduler with an initial learning rate of 0.01 and a Tmax of 200. 

For all other architectures and datasets, the optimizer used was SGD, momentum set to 0.9, with a weight decay of 1e-4, except for ResNet50 with TinyImagenet, which had a weight decay 5e-4. We used a cosine annealing scheduler with a starting learning rate of 0.01 and a Tmax of 160.  

\subsubsection{Adversarial attacks and adversarial training.}
Foolbox was used to implement PGD, FGSM, and AutoAttack for the adversarial attacks, with the PGD attack running 40 iterations. The adversarial training on ImageNet was performed using a 1-step PGD attack with an $\epsilon$ value of $4/255$\footnote{\url{https://github.com/dedeswim/vits-robustness-torch}}.

\subsection{Hardware Used:}
\label{sec:hardware}
We have used NVIDIA's two RTX 3090 GPUs and three A100 GPUs to compute all the simulations listed in the main paper and in the supplementary section.

\end{document}